\documentclass[10pt,twocolumn,letterpaper]{article}

\usepackage[pagenumbers]{cvpr}

\newcommand{\todo}[1]{}
\newcommand{\TODO}[1]{}

\definecolor{cvprblue}{rgb}{0.21,0.49,0.74}
\usepackage[pagebackref,breaklinks,colorlinks,allcolors=cvprblue]{hyperref}
\usepackage{subcaption}
\usepackage{cuted}
\usepackage[font=small]{caption}

\def\paperID{4076}
\def\confName{CVPR}
\def\confYear{2026}

\title{MambaEye: A Size-Agnostic Visual Encoder with Causal Sequential Processing}

\author{Changho Choi\\
Korea University\\
{\tt\small changho98@korea.ac.kr}
\and
Minho Luke Kim\\
MIT\\
{\tt\small lgmkim@mit.edu}
\and
Jinkyu Kim\textsuperscript{\dag}\\
Korea University\\
{\tt\small jinkyukim@korea.ac.kr}
}

\begin{document}
\maketitle
{\let\thefootnote\relax\footnotetext{$^{\dagger}$ Corresponding author}}
{\let\thefootnote\relax\footnotetext{Project page: \url{https://usingcolor.github.io/MambaEye}}}

\begin{abstract}
Despite decades of progress, 
a truly input-size agnostic visual encoder—a fundamental characteristic of human vision—has remained elusive. We address this limitation by proposing \textbf{MambaEye}, a novel, causal sequential encoder that leverages the low complexity and causal-process based pure Mamba2 backbone. Unlike previous Mamba-based vision encoders that often employ bidirectional processing, our strictly unidirectional approach preserves the inherent causality of State Space Models, enabling the model to generate a prediction at any point in its input sequence. A core innovation is our use of relative move embedding, which encodes the spatial shift between consecutive patches, providing a strong inductive bias for translation invariance and making the model inherently adaptable to arbitrary image resolutions and scanning patterns. To achieve this, we introduce a novel diffusion-inspired loss function that provides dense, step-wise supervision, training the model to build confidence as it gathers more visual evidence. We demonstrate that MambaEye exhibits robust performance across a wide range of image resolutions,
especially at higher resolutions such as $1536^2$ 
on the ImageNet-1K classification task. This feat is achieved while maintaining linear time and memory complexity relative to the number of patches.
\end{abstract}

\begin{figure}[t]
    \includegraphics[width=0.48\textwidth]{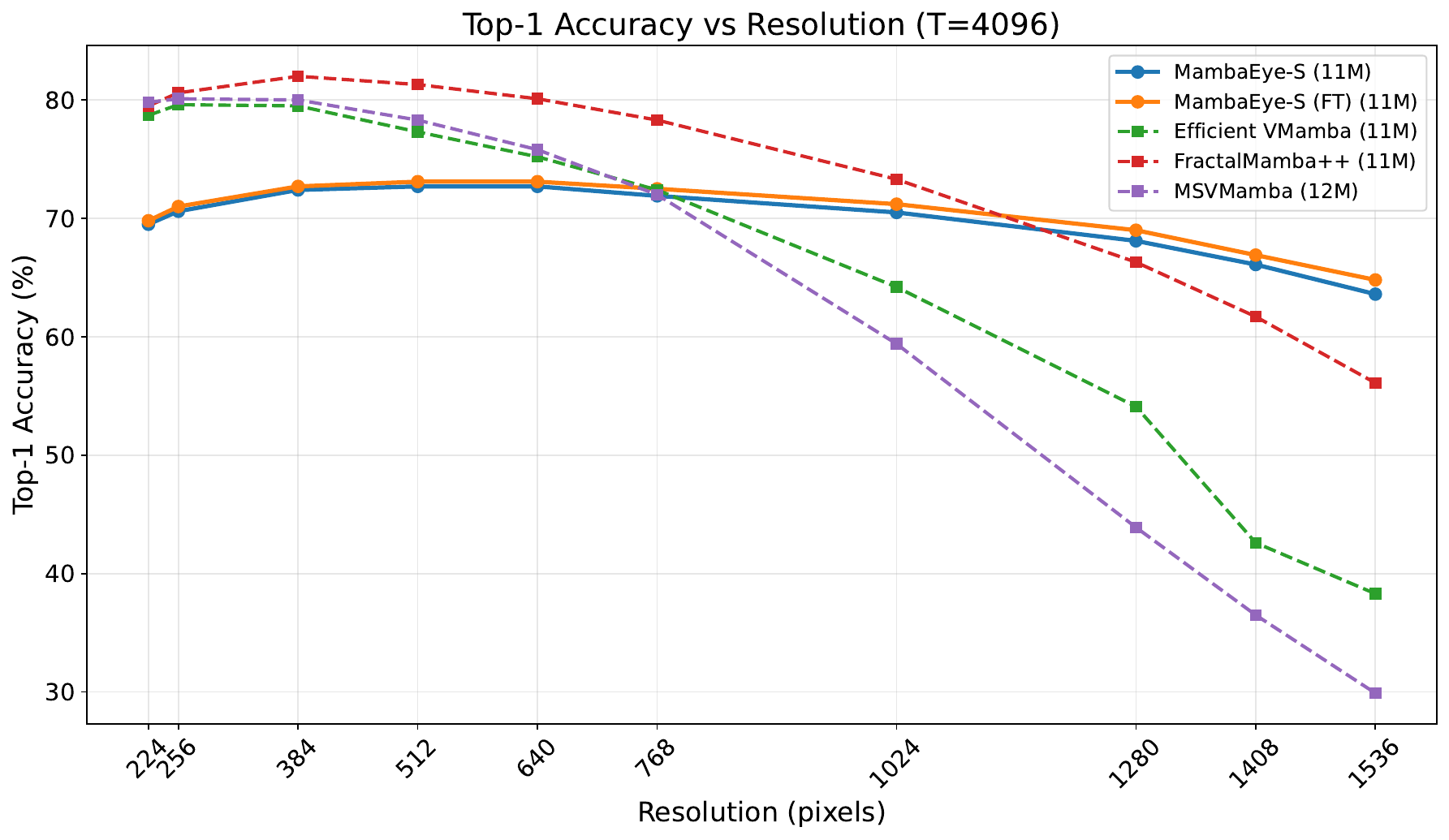}
    \caption{\textbf{MambaEye Resolution Scaling} Our MambaEye-S (11M params) models are benchmarked against Mamba-based models of similar size. While deterministic scanning methods like FractalMamba++~\cite{li2025scaling} achieve higher peak accuracy at medium resolutions, our model demonstrates superior scaling at extreme resolutions. Notably, MambaEye outperforms FractalMamba++ at resolutions over $1280^2$, despite using only a naive random sampling policy. This highlights our architecture's inherent robustness and size-agnostic capabilities.}
    \label{fig:teaser}
\end{figure}
\section{Introduction}
\label{sec:intro}

A central goal in computer vision is a truly \emph{input-size agnostic} encoder. These encoders will ideally operate on arbitrary resolutions and allocates resources based on availability, without committing to a fixed input size. While high resolution media is becoming more common, most vision encoders are limited on their ability to work on various input dimensions. Instead, these encoders typically resize or crop during the data piping process, incurring unavoidable information loss which could be detrimental to the task.

The disadvantages from this lack of versatility is already well-known in many models but is difficult to solve. Convolutional Neural Networks (CNNs)~\cite{lecun2002gradient, he2016deep}, while endowed with strong inductive biases, rely on fixed-size kernels whose effective receptive field grows slowly with depth, making global reasoning at high resolution costly. Vision Transformers (ViTs)~\cite{dosovitskiy2020image} provide global receptive fields but incur quadratic complexity in the number of patches, which rapidly becomes prohibitive for native-resolution inputs~\cite{prisadnikov2025vision}. Previous remedies such as linear/sparse attention~\cite{yang2024ralv, li2025polformer} reduce the cost at the expense of expressivity. To complicate matters, standard ViT pipelines require resizing and often overfits to the image size of the training data by learning the absolute positional embeddings. Even with sophisticated training strategies such as FixRes~\cite{touvron2019fixing} or relative encodings (e.g., RoPE)~\cite{su2021roformer, heo2024rotary}, the resolution brittleness persists.

State Space Models (SSMs) such as Mamba~\cite{gu2024mamba}, offer a better chance of realizing an input-size agnostic encoder due to their linear-time scaling and constant-memory recurrent inference, making them attractive for long sequences. However, when adapted to vision, most approaches serialize images with fixed, hand-crafted scanning paths~\cite{ibrahim2025survey} (e.g., raster scanning), disrupting 2D locality. Many Mamba based encoders also rely on bidirectional processing to compensate~\cite{liu2024vmamba, zhu2024vision}, which forfeits the memory advantages of causal recurrence. Recent work on fractal/space-filling traversals improves locality~\cite{li2025scaling, xiao2025boosting} but still enforces a predefined, holistic scan, which is not even guaranteed to be a optimal path~\cite{kutscher2025reordering} and does not eliminate the need to pick a fixed resolution up front. 

To have a truly size-agnostic model, we can once again turn to human vision perception. A human gathers information by dynamically scanning a scene, not by processing with a certain deterministic path. From this, unlike other architectures, we emphasize the importance of causal and sequential nature of vision perception. An ideal model should see the same part of the image multiple times, without a dependence in the order they are seen. Human vision rarely concludes the scene in one shot; instead, it cumulates information through time to build a confident understanding.

With these motivations in mind, we propose \textbf{MambaEye}, a causal and sequential visual encoder built on from pure Mamba2 backbone~\cite{dao2024transformers} that directly addresses these issues. Instead of committing to a global, fixed traversal or training resolution, MambaEye processes an \emph{arbitrary} sequence of patches and can emit a prediction at any step. Two design choices enable this:
(1) a \textbf{relative move embedding} that encodes the spatial displacement between consecutive patches, providing a strong inductive bias for translation invariance and making the model independent of image resolution and scan order; and
(2) a \textbf{diffusion-inspired loss}~\cite{ho2020denoising, bansal2023cold}, conceptually related to flow matching~\cite{lipman2023flow}, that provides dense, step-wise supervision by scheduling targets from a uniform prior to the ground truth according to observed information.

This design preserves the causal, recurrent nature of SSMs at inference, enabling constant memory per step and predictions at any point in the sequence.

Interestingly, this can be achieved without the need of deterministic scanning patterns. For MambaEye, random sampling is much more successful compared to more sophisticated scanning protocols such as raster scan~(Table~\ref{tab:scan_ablation}).

From this new paradigm, an additional point of improvement is introducing a new diffusion-inspired loss. This loss function guides the model to calibrate its confidence at any arbitrary timestep, which leads to improved generalization on longer sequences. This also incorporates the human vision perception where we cumulate information through time.
Indeed, compared to the standard cross entropy loss (Table~\ref{tab:loss_ablation}). 

Empirically, MambaEye achieves a competitive ImageNet-1K~\cite{imagenet15russakovsky} results while exhibiting robust scaling across resolutions and outperforming 
previous Mamba based encoders,
such as FractalMamba++~\cite{li2025scaling} at larger image sizes~(Figure~\ref{fig:teaser}).

Our primary contributions can be summarized as the following:
\begin{enumerate}
    \item We propose \textbf{MambaEye}, a novel and efficient visual encoder that is size-agnostic and causal.
    \item We introduce a \textbf{relative move embedding} technique that provides translation invariance and flexibility for arbitrary scanning paths.
    \item We design a new \textbf{diffusion-inspired loss function} that provides dense, step-wise supervision for sequential encoders.
\end{enumerate}

\section{Related Work}

\subsection{Sequential and Active Vision Paradigms}

Sequential processing of visual information has long been studied as a biology-inspired alternative to holistic image encoding. Early work in recurrent attention models demonstrated that neural networks could learn to selectively attend to image regions through reinforcement learning, processing high-resolution images via sequences of focused glimpses~\cite{mnih2014recurrent,ba2014multiple}. This paradigm is inspired by the saccadic movements of human eyes, in that rapid shift of attention across a visual scene rather than processing all information simultaneously is how humans perceive information.

Recent advances have extended these ideas in several directions. The Continuous Thought Machine~\cite{darlow2025continuous} explores how neural dynamics and synchronization can enable emergent reasoning capabilities, using sparse supervision at selected timesteps rather than dense feedback. Other approaches have investigated information-theoretic frameworks for modeling eye movements~\cite{renninger2005information} and generative models that learn task-driven sequential processing strategies~\cite{parr2021generative}. These works demonstrate that sequential visual processing can be both efficient and effective when properly supervised. This paradigm is now being scaled into general-purpose Active-Vision Foundation Models, such as CanViT~\cite{berreby2026canvit}, which successfully process arbitrary sequences of localized glimpses to perform complex dense predictions like semantic segmentation.

\subsection{Network Architectures in Vision Encoders}
Vision encoders have evolved from Convolutional Neural Networks (CNNs), which leverage locality and translation equivariance for hierarchical feature learning but struggle with global dependencies due to fixed kernels~\cite{he2016deep,lecun2002gradient}. 

Vision Transformers (ViTs) tackled this issue by incorporating self-attention to image patches, achieving global receptive fields early on. This however has come with the cost of quadratic complexity in patch count and heavy data requirements~\cite{dosovitskiy2020image}.

SSMs have recently mitigated these trade-offs in computer vision. Models like Vision Mamba (ViM)~\cite{zhu2024vision} and its variants adapt bidirectional SSMs for linear-time processing of flattened image sequences, demonstrating strong performance in classification and beyond~\cite{liu2024vmamba}. 

Hybrid approaches such as MambaVision~\cite{hatamizadeh2025mambavision} combine Mamba with Transformer elements for enhanced expressivity. Other variants, including VL-Mamba~\cite{qiao2024vlmamba} for multimodal learning and GroupMamba~\cite{shaker2025groupmamba} for vision tasks with efficient time and memory complexity, further expand the applicability of Mamba-based architectures. 

Recent surveys~\cite{zhang2024survey,liu2024visionma} highlight Mamba's growing adoption in new domains including remote sensing and image restoration, with scaling across resolutions. However, these models still treat images as static 1D sequences via raster scans, lacking dynamical, biologically inspired processing.

\section{Method}
\subsection{Preliminaries}
\paragraph{State Space Models.} Our model is built upon a State Space Model (SSM) backbone. SSMs map a 1D input sequence $x(t)$\footnote{In our model, the input sequence will be the sequence of image patches randomly extracted from the input image, which we will discuss later.} to an output $y(t)$ through a latent state $h(t)$, governed by a set of linear Ordinary Differential Equations (ODEs):
\begin{align}
    \frac{dh(t)}{dt} &= \mathbf{A}h(t) + \mathbf{B}x(t) \\
    y(t) &= \mathbf{C}h(t)
\end{align}
where $\mathbf{A}, \mathbf{B}, \mathbf{C}$ are learned state matrices. This set of ODEs can be discretized using a timescale parameter $\Delta$, yielding a recurrent formula:
\begin{equation}
    h_t = \bar{\mathbf{A}}h_{t-1} + \bar{\mathbf{B}}x_t, \quad y_t = \mathbf{C}h_t
\end{equation}
The Mamba architecture~\cite{gu2024mamba, dao2024transformers} significantly advanced SSMs by introducing a selection mechanism that makes the model's parameters input-dependent, dramatically improving its ability to selectively process information. A key advantage of SSMs is their computational efficiency. During inference, they can operate as a recurrent system with constant memory per step; during training, it acts as a parallelizable convolutional system with linear-time complexity in sequence length.

\paragraph{Sequential Problem Formulation.} Our work frames image classification as a sequential decision-making process. The model takes an input image $\mathbf{I} \in \mathbb{R}^{C \times H \times W}$, where $C$ is the number of channels and $H$ and $W$ are the image height and width. The height and width are not constrained to a fixed size. From this image, it processes a sequence of inputs over $T$ discrete time steps, $t \in \{0, ..., T-1\}$. The model's final output is a sequence of classification logits $\mathbf{Y} \in \mathbb{R}^{T \times N}$, where $N$ is the number of classes. Due to the model's causal nature, each vector $\mathbf{y}_t\in\mathbb{R}^{N}$ in the sequence represents the prediction after observing the first $t$ inputs.

\subsection{Data Processing}
To create a model that mimics human-like visual processing, we reconsidered the standard data preprocessing pipeline. Our approach is based on sequential recognition, allowing the model's capabilities to be independent of a fixed input resolution. To implement this, we extract a sequence of $P \times P$ image patches from arbitrary coordinates, selected uniformly at random during training. This mirrors the saccadic movement of human eyes and contrasts with the fixed, non-overlapping grid used by ViTs~\cite{dosovitskiy2020image}.

To ensure robustness to various image resolutions, augmented images are not resized but are instead pasted onto the center of a larger, zero-padded canvas.
For patch extraction, we sample from either exclusively from the original image region or from the entire canvas including the padding with equal probability. We also incorporate standard augmentation techniques such as random crops, perspective transformations, and color jittering to our training set.

\begin{figure}[h]
    \centering
    \includegraphics[width=0.48\textwidth]{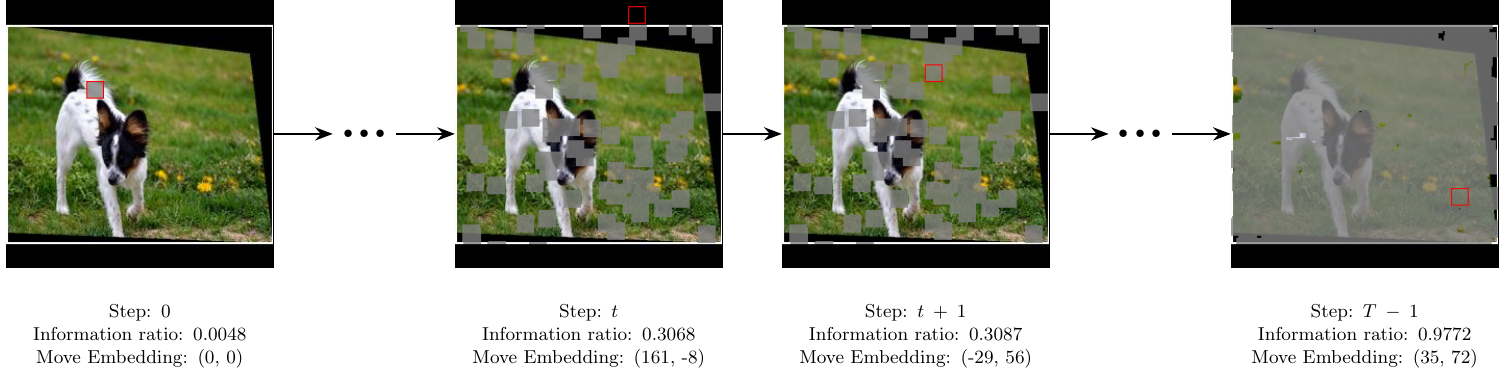}

    \caption{\textbf{Illustration of the Sequential Data Processing Pipeline.} The figure shows four snapshots of the data generation process. An original image is augmented and placed on a padded canvas (outlined by the white frame). At each step $t$, a new $16 \times 16$ patch (red border) is randomly sampled. The cumulative area covered by all patches is colored gray, and its proportion to the total image area defines the information ratio ($r_t$). A move embedding, $\mathbf{m}_t$, is also computed from the relative coordinate shift from the previous patch, providing spatial context to the model.}
    \label{fig:preprocess}
\end{figure}

For each step $t$, the model generates three key components as the input. This process is visualized in Figure~\ref{fig:preprocess}:
\begin{itemize}
    \item \textbf{Image Patch Vector ($\mathbf{v}_t$):} A flattened $P \times P$ image patch, where $\mathbf{v}_t \in \mathbb{R}^{d_{\text{image}}}$ and $d_{\text{image}} = C \times P \times P$.
    \item \textbf{Move Embedding ($\mathbf{m}_t$):} A sinusoidal position encoding of the relative coordinate change $(\Delta x, \Delta y)$ from the previous patch. The components are concatenated to form the move embedding, $\mathbf{m}_t \in \mathbb{R}^{d_{\text{move\_emb}}}$. By encoding relative, rather than absolute positions, this approach provides a crucial inductive bias for translation invariance and is the key to the model's ability to operate on images of arbitrary shape and resolution. The initial embedding, $\mathbf{m}_0$, is a zero vector.

    \item \textbf{Information Ratio ($r_t$):} A scalar value, $r_t \in [0, 1]$, measuring the fraction of the image's unique area covered by the cumulative set of patches up to step $t$. This ratio is crucial for guiding the training loss.
\end{itemize}

\subsection{Model Architecture}
\label{sec:model_arch}
\begin{figure}[h]
    \centering
    \includegraphics[width=0.45\textwidth]{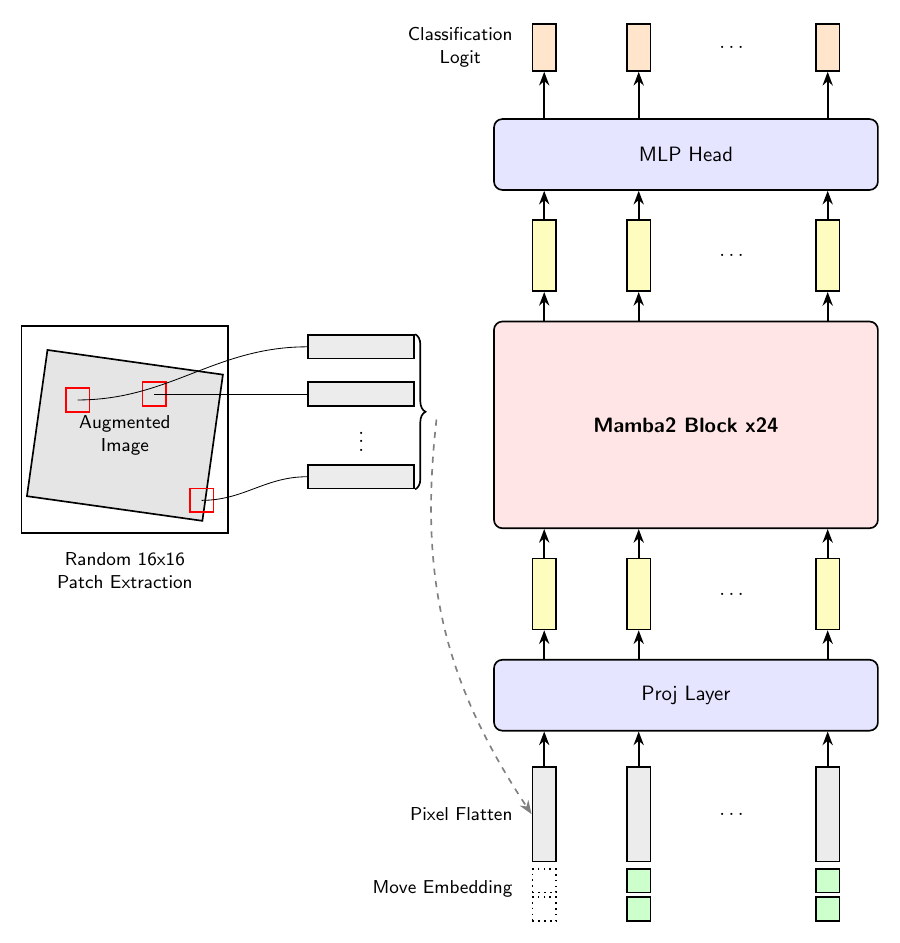}

    \caption{\textbf{MambaEye Model Architecture.} The diagram illustrates the sequential flow of the model. At each step, a flattened image patch vector and a sinusoidal "Move Embedding" are concatenated and processed by a projection layer. The core of the model is a stack of 24 Mamba2 blocks, whose output is passed to a final MLP head to produce a classification logit for that step.}
    \label{fig:model_arch}
\end{figure}

MambaEye processes a sequence of image patches to perform classification. The architecture consists of three main components, as illustrated in Figure~\ref{fig:model_arch}.

\paragraph{Projection Head.} At each step $t$, the image patch vector $\mathbf{v}_t$ and the move embedding $\mathbf{m}_t$ are concatenated to form an input vector $\mathbf{x}_t = [\mathbf{v}_t; \mathbf{m}_t] \in \mathbb{R}^{d_{\text{input}}}$, where $d_{\text{input}} = d_{\text{image}} + d_{\text{move\_emb}}$. This vector is then projected into the model's hidden dimension, $d_{\text{model}}$, by a single-layer MLP, denoted $f_{\text{proj}}$, with a GELU activation~\cite{hendrycks2016gaussian}.
\begin{equation}
    \mathbf{z}^{0}_{t} = f_{\text{proj}}(\mathbf{x}_t)
\end{equation}

\paragraph{Mamba2 Backbone.} The Mamba2 backbone is composed of $L$ stacked blocks. Inherited from the underlying SSM, each block $f_{\text{mamba}}^{i}$ supports dual inference modes. During training, it operates in a parallel mode, processing the entire sequence of embeddings $\mathbf{Z}^{i}$ at once using an efficient convolutional representation:
\begin{equation}
    \mathbf{Z}^{i+1} = f_{\text{mamba, parallel}}^{i}(\mathbf{Z}^{i}) \quad \text{for } i \in \{0, ..., L-1\}
\end{equation}
For inference, it can operate in a recurrent mode, processing one token at a time by updating a hidden state $\mathbf{h}_t^i$:
\begin{equation}
    (\mathbf{z}_{t}^{i+1}, \mathbf{h}_{t}^{i+1}) = f_{\text{mamba, recurrent}}^{i}(\mathbf{z}_{t}^{i}, \mathbf{h}_{t-1}^{i+1})
\end{equation}
This recurrent formulation enables efficient, step-by-step processing with constant memory. Our choice of a unidirectional, causal flow—in contrast to the bidirectional approach often used in other vision backbones—is a deliberate design choice to preserve this structure, which is fundamental to our sequential framework.

\paragraph{Classification Head.} Finally, each output representation $\mathbf{z}^{L}_t$ from the backbone is passed through a single-layer MLP, the classification head $f_{\text{head}}$, to produce the classification logits for all $N$ classes.
\begin{equation}
    \mathbf{y}_{t} = f_{\text{head}}(\mathbf{z}^{L}_{t})
\end{equation}
This process yields the full sequence of predictions, $\mathbf{Y} = (\mathbf{y}_0, ..., \mathbf{y}_{T-1})$.

\begin{table*}[t] 
\centering
\caption{Efficiency and scaling analysis of MambaEye. All metrics are measured on a single H100 GPU. Throughput and parallel memory are measured with a batch size of 128. Recurrent mode latency and memory are measured with a batch size of 1.}
\label{tab:comp_efficiency}
\footnotesize
\begin{tabular}{@{}lcccccccc@{}}
\toprule
& \multicolumn{2}{c}{FLOPs (G)} & \multicolumn{2}{c}{Throughput (img/s) [Parallel]} & \multicolumn{1}{c}{Latency (ms/step) [Recurrent]} & \multicolumn{3}{c}{Peak Memory (GB)} \\
\cmidrule(lr){2-3} \cmidrule(lr){4-5} \cmidrule(lr){6-6} \cmidrule(lr){7-9}
Model & $T=1024$ & $T=4096$ & $T=1024$ & $T=4096$ & (Any $T$) & $T=1024$ [Par.] & $T=4096$ [Par.] & (Any $T$) [Rec.] \\
\midrule
MambaEye-T & 11.71 & 46.84 & 4373.28 & 1124.99 & 6.88 & 3.11 & 12.21 & 0.06 \\
MambaEye-S & 22.23 & 88.92 & 2229.14 & 573.84 & 13.22 & 3.13 & 12.23 & 0.08 \\
MambaEye-B & 43.27 & 173.08 & 1130.29 & 290.37 & 21.88 & 3.17 & 12.27 & 0.13 \\
\bottomrule
\end{tabular}
\end{table*}
\subsection{Diffusion-Inspired Loss Function}
Our loss function is designed to train the model via a curriculum inspired by the iterative refinement process of diffusion models~\cite{ho2020denoising, bansal2023cold} and flow matching models~\cite{lipman2023flow}. We frame classification as a sequential inference task, where the model's predictive distribution should evolve from a state of maximum uncertainty (a uniform prior) to one of complete certainty (the ground-truth label) as it gathers more visual information. 

Let $T$ be the sequence length and $N$ be the number of classes. Let $\mathbf{y}_t$ be the model's logit vector at step $t$. Let $\mathbf{p}_{\text{target}} \in \{0, 1\}^N$ be the one-hot encoded ground-truth label, and let $\mathbf{p}_{\text{prior}} \in \mathbb{R}^N$ be a uniform distribution where each element is $1/N$.

The schedule for this curriculum is governed by the information ratio, $r_t \in [0, 1]$. The scheduled target distribution at step $t$, denoted $\mathbf{p}_{\text{scheduled}}^{(t)}$, is defined as a linear interpolation:
\begin{equation}
    \mathbf{p}_{\text{scheduled}}^{(t)} = (1 - r_t) \cdot \mathbf{p}_{\text{prior}} + r_t \cdot \mathbf{p}_{\text{target}}
\end{equation}
The total loss $\mathcal{L}$ is the mean Cross-Entropy (CE) loss computed over the entire sequence between the model's logits and the corresponding scheduled target:
\begin{equation}
    \mathcal{L} = \frac{1}{T} \sum_{t=0}^{T-1} \text{CE}\left(\mathbf{p}_{\text{scheduled}}^{(t)}, \mathbf{y}_t\right)
\end{equation}
where the CE loss for a single step is given by:
\begin{equation}
    \text{CE}(\mathbf{p}, \mathbf{y}) = - \sum_{n=1}^{N} \mathbf{p}_n \log\left(\frac{\exp(\mathbf{y}_n)}{\sum_{i=1}^{N} \exp(\mathbf{y}_i)}\right)
\end{equation}
This formulation provides dense supervision at every step, explicitly guiding the model to calibrate its confidence with the amount of visual evidence it has processed. By requiring an accurate prediction at every point along the schedule, this loss also implicitly encourages the model's hidden state to act as a compressed summary of the visual information gathered, promoting efficient state management throughout the sequence.

\section{Results}
\begin{figure*}[t]
    \centering 
    \begin{subfigure}[b]{0.48\textwidth}
        \centering
        \includegraphics[width=\linewidth]{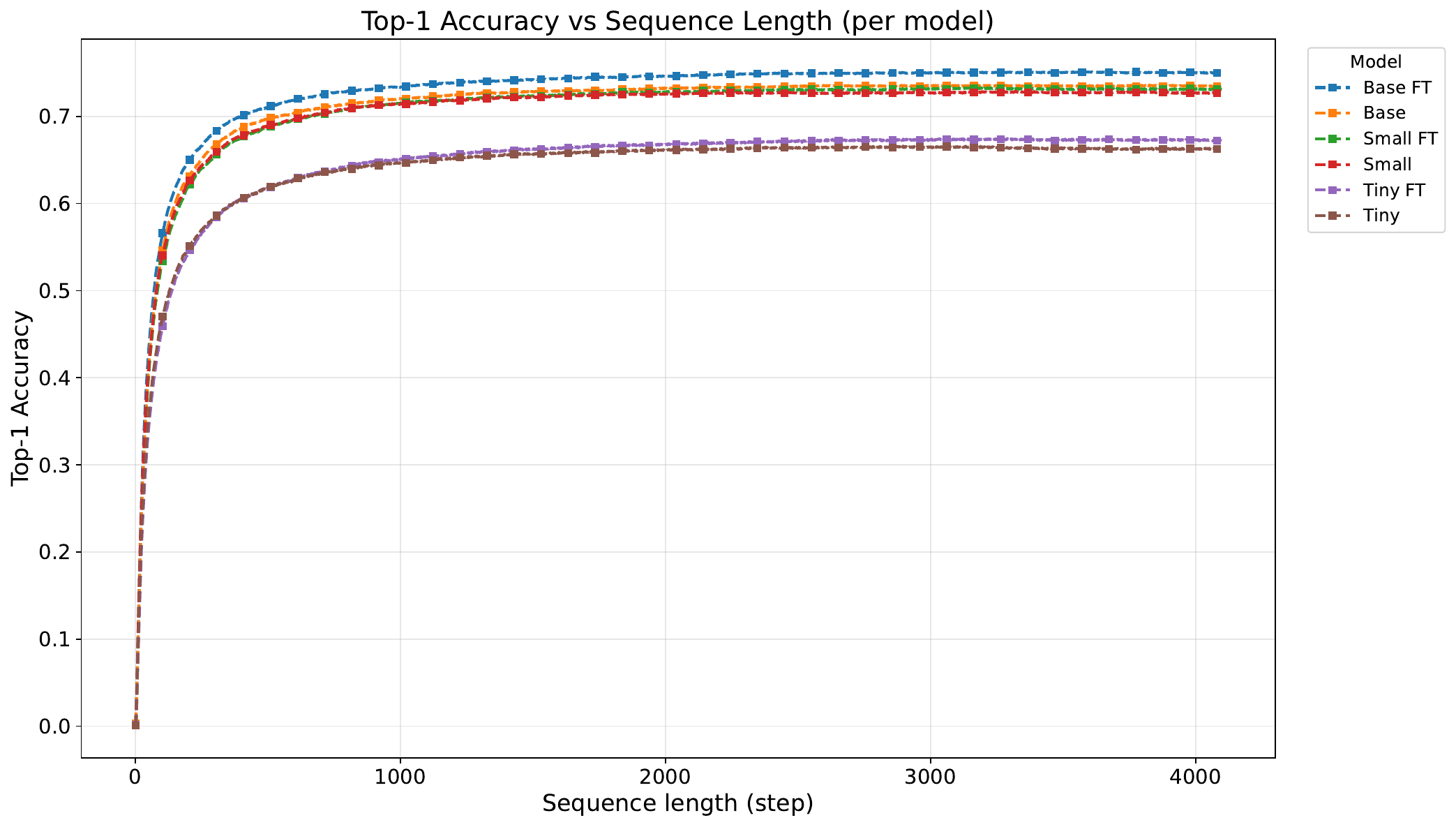}
       
        \caption{All MambaEye models at $512^2$
        }
        \label{fig:image_a}
    \end{subfigure}
    \hfill
    \begin{subfigure}[b]{0.48\textwidth}
        \centering
        \includegraphics[width=\linewidth]{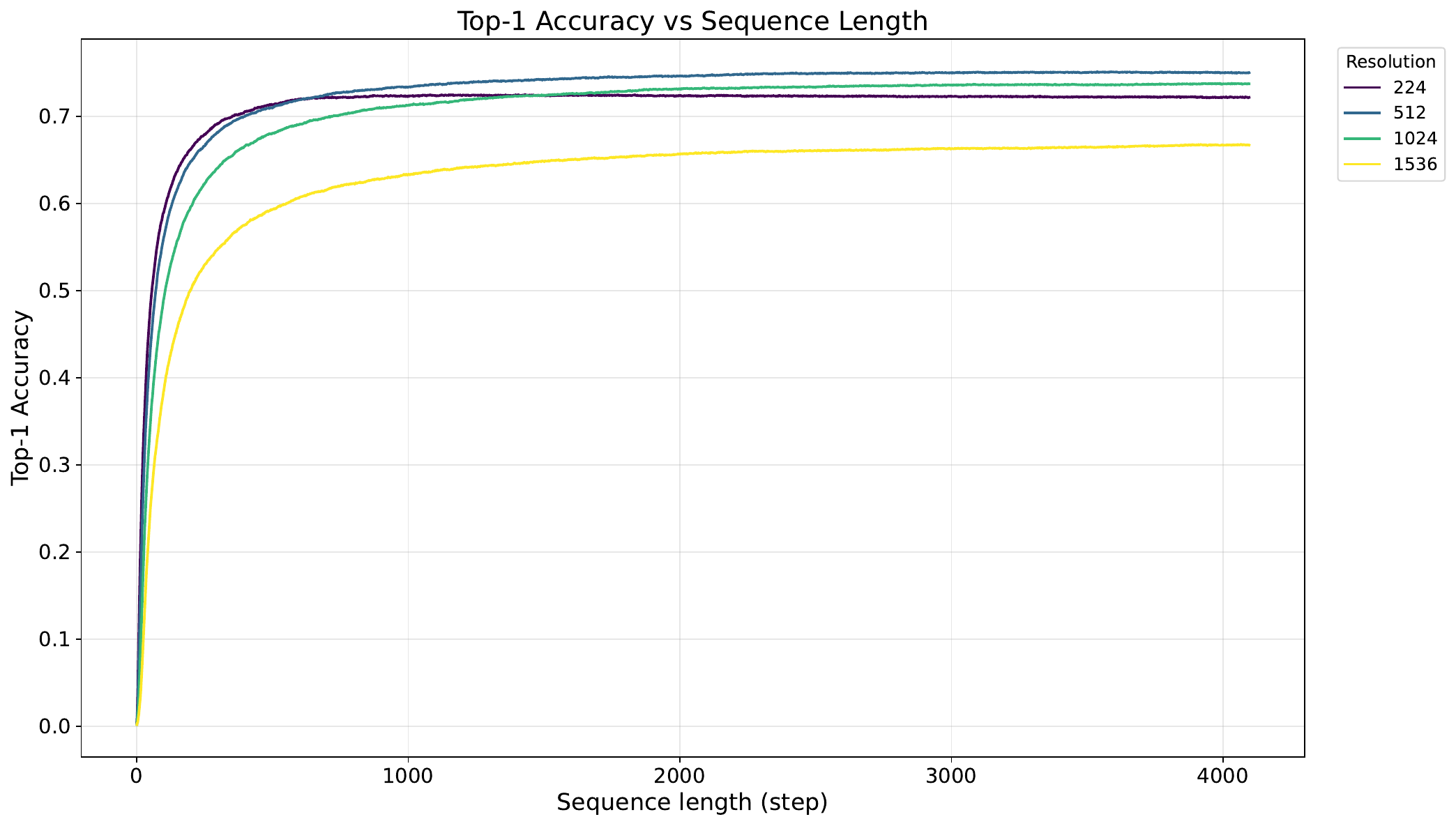}
        \caption{
        MambaEye-B (FT) at $224^2$, $512^2$, $1024^2$and $1536^2$
        }
        \label{fig:image_b}
    \end{subfigure}
    \caption{\textbf{MambaEye performance vs. sequence length ($T$).} (a) shows all model variants accumulating information at a fixed $512^2$ resolution. Accuracy saturates for all models, with fine-tuned models achieving higher performance. (b) shows the MambaEye-B (FT) model across different resolutions, illustrating the trade-off between image size and required sequence length. At lower resolutions, performance saturates quickly, while higher resolutions require more steps.}
    \label{fig:experiment_main}
\end{figure*}
\subsection{Implementation Details}
\paragraph{Training Setup.} We train our models from scratch for 300 epochs using the AdamW optimizer~\cite{loshchilov2017decoupled} with a total batch size of 1024. The learning rate follows a cosine decay schedule, warm up for 20 epochs to a peak of $1 \times 10^{-3}$ before decaying to zero. We use optimizer betas of $(0.9, 0.999)$ and a weight decay of $0.05$.

\paragraph{Data Parameters.} For data processing, we use a patch size of $P=16$, resulting in a patch vector dimension $d_{\text{image}} = 768$ (for 3-channel images). The move embedding has a dimension of $d_{\text{move\_emb}} = 512$, with 256 dimensions allocated to each of the x and y coordinates. This yields a total input dimension $d_{\text{input}} = 1280$. During pre-training, the canvas size for each image is independently sampled from a uniform random distribution between $224^{2}$ and $1024^{2}$ resolution. The pre-training sequence length is set to $T=1024$.

\paragraph{Model Configurations.} We experiment with three model sizes by varying the number of layers ($L$) while keeping the hidden dimension fixed to $d_{\text{model}} = 256$. Detailed configurations and model parameter sizes are tabulated at Table~\ref{tab:model_config} below.
\begin{table}[!h]
\centering
\caption{Model Configurations}
\begin{tabular}{@{}lcc@{}} 
\toprule
Model Config & Layers ($L$) & Parameters\\
\midrule
MambaEye-T (Tiny) & 12 & 5.8M\\
MambaEye-S (Small) & 24 & 11.0M\\
MambaEye-B (Base) & 48 & 21.3M\\
\bottomrule
\end{tabular}
\label{tab:model_config}
\end{table}
\paragraph{Fine-tuning.} Following pre-training, we fine-tune the models on a longer sequence length of $T=2048$ for 30 epochs. For this stage, we follow the protocol of Vision Mamba~\cite{liu2024vmamba}, using a reduced constant learning rate of $1 \times 10^{-5}$ and weight decay of $1 \times 10^{-8}$.

\subsection{Computational Efficiency}
We analyze the computational performance of our MambaEye models, focusing on the key advantages of its architecture. As discussed in Section~\ref{sec:model_arch}, our model supports dual inference modes: a parallel mode for high-throughput training and a recurrent mode for memory-efficient inference.

Table~\ref{tab:comp_efficiency} provides a detailed scaling analysis of our three model sizes. We report theoretical FLOPs and practical inference metrics for both modes across different sequence lengths. All practical metrics (throughput, latency, memory) are measured on a single NVIDIA H100 GPU using FP32 precision. The total FLOPs are identical for both modes at a given sequence length $T$ and FLOPs value is proportional to the sequence length. The recurrent mode's latency is reported per-step, and its memory footprint is constant, demonstrating the model's size-agnostic capability in memory complexity.

\begin{table*}[t!]
\centering
\caption{Top-1 Accuracy (\%) on ImageNet-1K at various resolutions ($T=4096$). ``FT'' denotes fine-tuning on $T=2048$; ``KD'' denotes knowledge distillation from DeiT-III-Base~\cite{touvron2022deit} at $T=1024$ (no long-sequence fine-tuning). Competitor results from Li et al.~\cite{li2025scaling}.}
\label{tab:size_ablation}

\begin{tabular}{@{}lcccccccccc@{}}
\toprule
Model & $224^{2}$ & $256^{2}$ & $384^{2}$ & $512^{2}$ & $640^{2}$ & $768^{2}$ & $1024^{2}$ & $1280^{2}$ & $1408^{2}$ & $1536^{2}$ \\
\midrule
\textbf{MambaEye-T} (5.8M) & 61.1 & 62.9 & 65.8 & 66.2 & 66.2 & 65.3 & 63.3 & 60.6 & 58.8 & 56.5 \\
\textbf{MambaEye-T (FT)} (5.8M) & 62.1 & 63.7 & 66.7 & 67.2 & 67.1 & 66.4 & 64.4 & \textbf{61.6} & \textbf{59.8} & \textbf{57.4} \\
\midrule
{MSVMamba (7M)}~\cite{shi2024multi}&    
\textbf{77.3} & {77.7} & {77.4} & {75.0} & {71.7} & {65.8} & {48.0} & {31.0}& {23.8} & {18.3}\\
{ViM (7M)}~\cite{zhu2024vision} & 
{76.1} & {76.3} & {70.4} & {67.4} & {51.4} & {30.6} & {16.1} & {7.2} & {4.1}& {1.8}\\
{Efficient VMamba (6M)}~\cite{pei2025efficientvmamba} &
{76.5} & {76.9 } & {76.5} & {73.8} & {70.4} & {65.8} & {52.0 } & {36.2}& {29.4} & {24.1}\\
{FractalMamba++ (7M)}~\cite{li2025scaling} &   
\textbf{77.3} & \textbf{78.4} & \textbf{79.5} & \textbf{78.4} & \textbf{76.4} & \textbf{73.7} & \textbf{66.5} & {55.2}& {48.1} & {42.5}\\
\midrule
\midrule
\textbf{MambaEye-S} (11M) & 69.5 & 70.6 & 72.4 & 72.7 & 72.7 & 71.9 & 70.5 & 68.1 & 66.1 & 63.6 \\
\textbf{MambaEye-S (FT)} (11M) & 69.8 & 71.0 & 72.7 & 73.1 & 73.1 & 72.5 & 71.2 & \textbf{69.0} & \textbf{66.9} & \textbf{64.8} \\
\midrule
{MSVMamba (12M)}~\cite{shi2024multi} & 
\textbf{79.8} & {80.1} & {80.0} & {78.3} & {75.8} & {72.0} & {59.4} & {43.9}& {36.5} & {29.9}\\
{Efficient VMamba (11M)}~\cite{pei2025efficientvmamba} & {78.7} & {79.6} & {79.5} & {77.3} & {75.2} & {72.4} & {64.2} & {54.1}& {42.6} & {38.3}\\
{FractalMamba++ (11M)}~\cite{li2025scaling} & {79.5} & \textbf{80.6} & \textbf{82.0} & \textbf{81.3} & \textbf{80.1} & \textbf{78.3} & \textbf{73.3} & {66.3}  & {61.7} & {56.1} \\
\midrule
\midrule
\textbf{MambaEye-B} (21M) & 70.6 & 71.7 & 73.3 & 73.5 & 73.4 & 72.9 & 71.7 & 69.4 & 67.0 & 64.3\\
\textbf{MambaEye-B (FT)} (21M) & 72.2 & 73.2 & 74.8 & 75.0 & 75.0 & 74.6 & 73.7 & 71.5 & 69.5 & 66.7 \\
\textbf{MambaEye-B (KD)} (21M) & 74.1 & 74.8 & 75.5 & 75.5 & 75.4 & 75.0 & 74.1 & 71.7 & 69.3 & 66.5 \\
\midrule
{VMamba (31M)}~\cite{zhu2024vision} & {82.5} & {82.5} & {82.5} & {81.1} & {79.3} & {76.1} & {62.3} & {50.2} & {45.1} & {40.9} \\
{FractalMamba++ (30M)}~\cite{li2025scaling} & \textbf{83.0} & \textbf{83.5} & \textbf{84.1} & \textbf{83.9} & \textbf{83.0} & \textbf{81.9} & \textbf{78.8} & \textbf{74.3} & \textbf{71.3} & \textbf{67.5} \\
\bottomrule
\end{tabular}
\end{table*}

\subsection{Training Results on ImageNet1K}
We evaluate all model configurations (with and without fine-tuning) on the ImageNet-1K validation set across a wide range of image resolutions. For inference, images are resized to the target resolution while preserving their original aspect ratio, and then placed on a zero-padded canvas. Crucially, all patch sampling is constrained to the original image area, not the padding.

A unique feature of MambaEye is its ability to produce a prediction at any sequence length. As shown in Figure~\ref{fig:experiment_main}, we can analyze performance as a function of $T$. For most evaluations, we standardize our main results at a sequence length of $T=4096$, as model performance typically saturates before this point. While performance generally increases with longer sequences, we note an interesting anomaly in Figure~\ref{fig:experiment_main}(b): at low resolutions (e.g., $224^{2}$ resolution), performance can degrade after a certain number of steps.

We present a series of controlled ablations to analyze our model's key characteristics, including the impact of our diffusion-inspired loss, the effect of image size, and the choice of scanning pattern.

\subsubsection{Loss Function Ablation}
To validate the effect of our proposed diffusion-inspired loss, we compare the two models with both using the MambaEye-T architecture, and without fine-tuning process, evaluated at $T=4096$. For comparison, one of the models is trained with a standard cross entropy (CE) loss, applied to the output logit of every sequence, $\mathbf{y}_{t}$,   $t \in \{0, ..., T-1\}$, while the other utilizes our new diffusion-based loss.

Table~\ref{tab:loss_ablation} demonstrates that our new loss function has a demonstrable impact on the performance. The baseline model trained with a standard cross entropy loss achieves a lower top-1 accuracy in every resolution.

This result strongly validates our core hypothesis: for a sequential processing model, dense, step-wise supervision is beneficial. The standard cross entropy loss is less optimized in providing sufficient training signals for the model to learn how to effectively cumulate and refine information over time. Our diffusion-inspired loss provides a clear framework, teaching the model to build a progressively more accurate prediction as it gathers more visual evidence. More details are provided in the Supplementary Material.

\begin{table}[!h]
\centering
\caption{Ablation study of the loss function on ImageNet-1K.}
\label{tab:loss_ablation}
\begin{tabular}{@{}lcccc@{}}
\toprule
Loss Function & $224^{2}$ & $512^{2}$ & $1024^{2}$ & $1536^{2}$ \\
\midrule
Standard CE & 56.7 & 61.9 & 59.9 & 53.6 \\
Ours (Diffusion-inspired) & 61.1 & 66.2 & 63.3 & 56.5 \\
\bottomrule
\end{tabular}
\end{table}

\subsubsection{Ablation on Image Size}
A key claim of our work is that MambaEye's architecture is size-agnostic. To test this, we evaluate our pretrained models, with and without long-sequence fine-tuning, on a wide range of image resolutions from $224^2$ to $1536^2$. For each evaluation, the model processes a sequence of $T=4096$ randomly sampled patches. The results are shown in Table~\ref{tab:size_ablation}. We compare our models against recent Mamba-based backbones designed for resolution scaling.

Across Tiny, Small, and Base, accuracy tends to peak around mid resolutions (roughly $384^2$–$640^2$) and then softens as resolution continues to grow. This is consistent with the fixed sequence budget: with $T{=}4096$, higher resolutions dilute the coverage per step, so the model needs more steps to get full information from the image. Scaling to larger models help consistently, the fine-tuned variants of Tiny/Small/Base reach 67.2/73.1/75.0 at $512^2$. The advantage of scaling persists even at $1536^2$, with 57.4/64.8/66.7 top-1 accuracy, but the overall shape of the curve remains similar. Fine-tuning also plays a clear role. It increases accuracy across the board and makes the high resolution tail more resilient (e.g., at $1536^2$, +0.9/+1.2/+2.4 points for Tiny/Small/Base).

The high resolution behavior is remarkably uniform across model sizes: the absolute drop from the peak to $1536^2$ is on the order of 8–10 points for all three scales (e.g., Small (FT) 73.1→64.8 and Base (FT) 75.0→66.7).

At low and mid resolutions, methods like FractalMamba++ typically attain higher peaks at similar parameter counts. However, as we move into the high-resolution regime, the gap narrows or reverses for smaller models: at $1280^2$ to $1536^2$, our Tiny and Small (FT) variants surpass size-matched baselines, and the Base (FT, 21M) model closes on 30–31M-parameter alternatives. This aligns with our central claim. Even with a naive random sampling policy, the architecture scales robustly with resolution.

\subsubsection{Ablation on Scanning Patterns}
To understand the impact of the scanning policy, we conducted a comprehensive evaluation of nine scanning patterns using MambaEye-B (FT), spanning both deterministic and stochastic strategies.

For deterministic patterns, we split the image into a grid of non-overlapping $16 \times 16$ patches, preserved the original aspect ratio on a canvas, and extracted patches only from the valid image area. Patterns include: \textbf{Column Raster} (top-to-bottom, left-to-right), \textbf{Horizontal Raster} (left-to-right, top-to-bottom), \textbf{Horizontal Snake} and \textbf{Column Snake} (boustrophedon row/column traversals), \textbf{Diagonal}, \textbf{Spiral} (outside-in), and \textbf{Hilbert} (space-filling curve)~\cite{kutscher2025reordering}. We also test a \textbf{Golden} scan: patches are visited by jumping $\approx 61.8\%$ of the total sequence length at each step (Golden Ratio stride), ensuring maximally spread-out sampling without randomness. Finally, \textbf{Random (fixed grid)} uses random ordering of non-overlapping grid patches, isolating the effect of patch overlap from scan order. As with other ablations, we use $T{=}4096$. For random scanning, we report the mean and standard deviation over 5 independent runs.

The results in Table~\ref{tab:scan_ablation} reveal a clear dichotomy. Patterns that traverse mostly \emph{adjacent} patches (Column Raster, Raster, Snake variants, Diagonal, Spiral, Hilbert) collapse at high resolutions because the model lacks early global context---the causal state is saturated with local information before any distant region is visited. In contrast, patterns that provide \emph{early global coverage}---Random and Golden---perform strongly across all resolutions. The Golden scan achieves this deterministically by maximally spacing consecutive visits, closely matching Random scanning at higher resolutions without any stochasticity. This demonstrates that the critical factor is not randomness per se, but whether the scanning policy exposes global context early in the sequence.

Random scanning with a fixed non-overlapping grid (no patch overlap) shows a significant accuracy drop at lower resolutions ($60.3\%$ vs.\ $73.5\%$ at $224^{2}$), demonstrating that MambaEye benefits from the flexible, redundant sampling enabled by our grid-free approach. At higher resolutions where coverage is sparse regardless, the gap narrows. Notably, random scanning is highly stable across seeds ($\sigma < 0.001$). Step-wise accuracy curves for deterministic patterns are provided in the Supplementary Material.

\begin{table}[h]
\centering
\caption{Scanning path ablation on ImageNet-1K using MambaEye-B (FT).}
\label{tab:scan_ablation}
\footnotesize
\setlength{\tabcolsep}{4pt}
\begin{tabular}{@{}lcccc@{}}
\toprule
Scanning Path & $224^{2}$ & $512^{2}$ & $1024^{2}$ & $1536^{2}$ \\
\midrule
\textbf{Random (mean)} & \textbf{73.5} & \textbf{74.9} & 73.7 & 67.0 \\
\textit{Random ($\sigma$)} & \textit{0.0003} & \textit{0.0004} & \textit{0.0007} & \textit{0.0007} \\
\midrule
Random (fixed grid) & 60.3 & 73.6 & \textbf{73.8} & 66.6 \\
Golden & 66.4 & 73.6 & 73.3 & \textbf{67.7} \\
\midrule
Diagonal & 51.2 & 51.4 & 26.0 & 9.4 \\
Spiral & 51.6 & 45.8 & 27.2 & 7.6 \\
H. Raster & 49.6 & 44.9 & 18.7 & 8.0 \\
Hilbert & 45.0 & 40.7 & 24.1 & 12.9 \\
Column Raster & 45.3 & 38.9 & 12.5 & 6.0 \\
H. Snake & 43.2 & 31.5 & 13.8 & 6.4 \\
C. Snake & 42.4 & 35.6 & 14.0 & 6.0 \\
\bottomrule
\end{tabular}
\end{table}

\subsubsection{Knowledge Distillation}
To investigate whether the accuracy gap between MambaEye and established baselines is architectural or a consequence of our vanilla training recipe, we conduct a knowledge distillation (KD) experiment. A pretrained DeiT-III-Base~\cite{touvron2022deit} (86M, 83.8\% top-1) serves as the teacher. The teacher processes each image once at $224^{2}$ and produces a logit vector $\mathbf{z}^{T}$. We augment our diffusion-inspired loss with a step-wise KL divergence term weighted by the information ratio $r_t$:
\begin{equation}
    \mathcal{L} = \alpha \, \mathcal{L}_{\text{diff}} + (1 - \alpha) \, \frac{1}{T} \sum_{t=0}^{T-1} r_t \cdot \tau^{2} \, \text{KL}\!\left(\sigma\!\left(\frac{\hat{\mathbf{y}}_{t}}{\tau}\right) \middle\| \sigma\!\left(\frac{\mathbf{z}^{T}}{\tau}\right)\right)
\end{equation}
where $\tau{=}4$ is the temperature, $\alpha{=}0.3$, and $\sigma$ denotes softmax. The $r_t$ weighting is critical: at early steps the student has seen few patches, so forcing it to match the teacher's full-image prediction would be counterproductive. We start from the pretrained MambaEye-B checkpoint and train for 30 epochs at $T{=}1024$ with a cosine-decayed learning rate of $5 \times 10^{-5}$. The teacher is frozen throughout.

As shown in Table~\ref{tab:size_ablation} (KD row), the distilled model surpasses both the pretrained and fine-tuned models from $224^{2}$ through $1024^{2}$, reaching $75.5\%$ at $512^{2}$ ($+2.0$ over Base, $+0.5$ over FT) without any long-sequence fine-tuning. Only at $1408^{2}$--$1536^{2}$ does it fall marginally behind the fine-tuned variant, which is expected since FT explicitly trains on $T{=}2048$. Notably, the teacher only sees $224^{2}$ images, yet the student benefits across all resolutions up to $1536^{2}$, confirming that size-agnosticism is preserved under distillation. These results indicate that a significant portion of the accuracy gap stems from the training signal rather than architectural limitations. Full step-wise accuracy curves are provided in the Supplementary Material.

\section{Conclusion}
In this work, we introduced \textbf{MambaEye}, a novel visual encoder that re-frames image recognition as a causal, sequential-processing task. Our approach is built on a pure Mamba2 backbone, preserving its linear-time efficiency and constant-memory inference. We presented two core innovations: a \textbf{relative move embedding} that provides translation invariance for arbitrary scanning paths, and a \textbf{diffusion-inspired loss function} that provides dense, step-wise supervision.

Our experimental results validate this new paradigm. We demonstrated that our diffusion-inspired loss is critical, improving performance 
compared to a standard cross entropy loss~(Table~\ref{tab:loss_ablation}). We also proved that our architecture's flexibility is genuine. The model's performance with a simple random scan dramatically exceeds deterministic scanning patterns (Table~\ref{tab:scan_ablation}), confirming that the relative move embedding successfully provides spatial context.

While our model's peak accuracy does not yet match architectures with deterministic scanning path at lower resolutions, we show that its performance scaling is highly robust, outperforming previous Mamba based encoders for small and intermediate sized models. Furthermore, knowledge distillation from a grid-based teacher (DeiT-III) closes much of the gap, reaching $75.5\%$ at $512^2$ without long-sequence fine-tuning, the results highlight that further performance improvements can be unlocked through enhanced training recipes.

\paragraph{Limitations and future work}

While MambaEye is promising, we identify several limitations that guide future work. Our current training omits advanced augmentations like CutMix~\cite{yun2019cutmix} or Mixup~\cite{zhang2017mixup}, and our architectural study is limited. A further ablation on image augmentation and bringing up the performance for larger model parameters will also be helpful. In a more bird's eye view, it will be worth it to understand MambaEye's unidirectional architecture and its general compatibility with other complex structures like H-Net~\cite{hwang2025dynamic} could yield further improvements. A key future direction is a systematic study of generalization, not only to longer sequences~\cite{ruiz2025understanding} but also to higher dimensional data like video or 3D volumes, which could be supported by adapting the move embedding with higher dimension. Finally, the most significant future step is to replace random patch sampling with a learned, adaptive scanning policy, bringing the model's efficiency closer to that of human vision.

{
    \small
    \bibliographystyle{ieeenat_fullname}
    \bibliography{main}
}

\clearpage
\setcounter{page}{1}
\twocolumn{
  \maketitlesupplementary
}

\begin{strip}
\section{Model Implementation}
\label{sec:suppl_arch}

We provide the simplified PyTorch-like pseudo-code of the MambaEye model to clarify the architectural details. The model consists of an initial projection layer, a Mamba2 backbone, and a classification head.

\begin{verbatim}
class MambaEye(nn.Module):
    def __init__(self, num_classes, input_dim, dim, depth, 
                 d_state, d_conv, expand):
        super().__init__()
        # Initial projection
        self.init_fc = nn.Sequential(
            nn.Linear(input_dim, dim), 
            nn.SiLU()
        )

        # Mamba2 backbone with pre-norm architecture
        self.mamba2_net = Mamba2backbone(
            d_model=dim,
            n_layer=depth,
            ssm_cfg={
                "layer": "Mamba2",
                "d_state": d_state, # 128
                "d_conv": d_conv,   # 4
                "expand": expand,   # 2
            },
            rms_norm=True,
            fused_add_norm=True,
            residual_in_fp32=True
        )
        
        # Classification head
        self.classification_head = nn.Sequential(
            nn.Linear(dim, num_classes),
        )

    def forward(self, img_sequence, move_embedding):
        # Concatenate inputs: (B, L, D_img) + (B, L, D_move)
        x = torch.cat((img_sequence, move_embedding), dim=-1)

        # Initial projection
        x = self.init_fc(x)

        # Forward through Mamba2 backbone
        x, hidden_states_list = self.mamba2_net(x)

        # Classification head
        classification_sequence = self.classification_head(x)

        return classification_sequence
\end{verbatim}

\section{Data Processing Pipeline}
\label{sec:suppl_data}

Our data processing pipeline mimics saccadic eye movements by sampling patches from random locations. We provide the core logic for patch extraction and move embedding below.

\subsection{Sequential Patch Sampling}
\begin{verbatim}
def get_item(self, idx):
    # 1. Create a zero-padded canvas
    canvas = torch.ones(3, canvas_size, canvas_size) * padding_value
    
    # 2. Place image on canvas (centered)
    canvas[:, row_offset:row_offset+H, col_offset:col_offset+W] = img
    
    # 3. Randomly sample coordinates for the sequence
    # Sample top-left coordinates (r, c) for each step
    row_coords = randint(row_offset, row_end, sequence_length)
    col_coords = randint(col_offset, col_end, sequence_length)
    
    # 4. Extract patches
    patches = []
    for i in range(sequence_length):
        r, c = row_coords[i], col_coords[i]
        patch = canvas[:, r:r+patch_size, c:c+patch_size]
        patches.append(patch.flatten())
    
    # 5. Compute relative moves
    # delta_r[t] = r[t] - r[t-1] (with r[-1] = 0)
    delta_r = cat([row_coords[0:1], row_coords[1:] - row_coords[:-1]])
    delta_c = cat([col_coords[0:1], col_coords[1:] - col_coords[:-1]])
    
    # 6. Encode moves
    move_embeddings = sinusoidal_enc_2d(
        stack([delta_c, delta_r], dim=1), move_dim // 2
    )
    
    return stack(patches), move_embeddings
\end{verbatim}

\section{Training Logic}
\label{sec:suppl_training}

The training step involves computing the diffusion-inspired loss, which compares the model's predicted logits against a scheduled target distribution.

\begin{verbatim}
def training_step(self, batch):
    img_sequence, move_embedding, info_ratio, labels = batch
    
    # Forward pass
    logits_seq = self.model(img_sequence, move_embedding)
    
    # Calculate Loss
    # p_target: One-hot ground truth
    # p_prior: Uniform distribution
    # p_scheduled = (1 - r_t) * p_prior + r_t * p_target
    
    loss = 0
    for t in range(T):
        r_t = info_ratio[t]
        target_dist = (1 - r_t) * uniform_prior + r_t * one_hot_label
        loss += cross_entropy(logits_seq[:, t], target_dist)
        
    return loss / T
\end{verbatim}
\end{strip}

\section{Additional Experimental Results}
\label{sec:suppl_experiments}

\subsection{Loss Ablation Analysis}
\begin{figure*}[h]
    \centering
    \includegraphics[width=\textwidth]{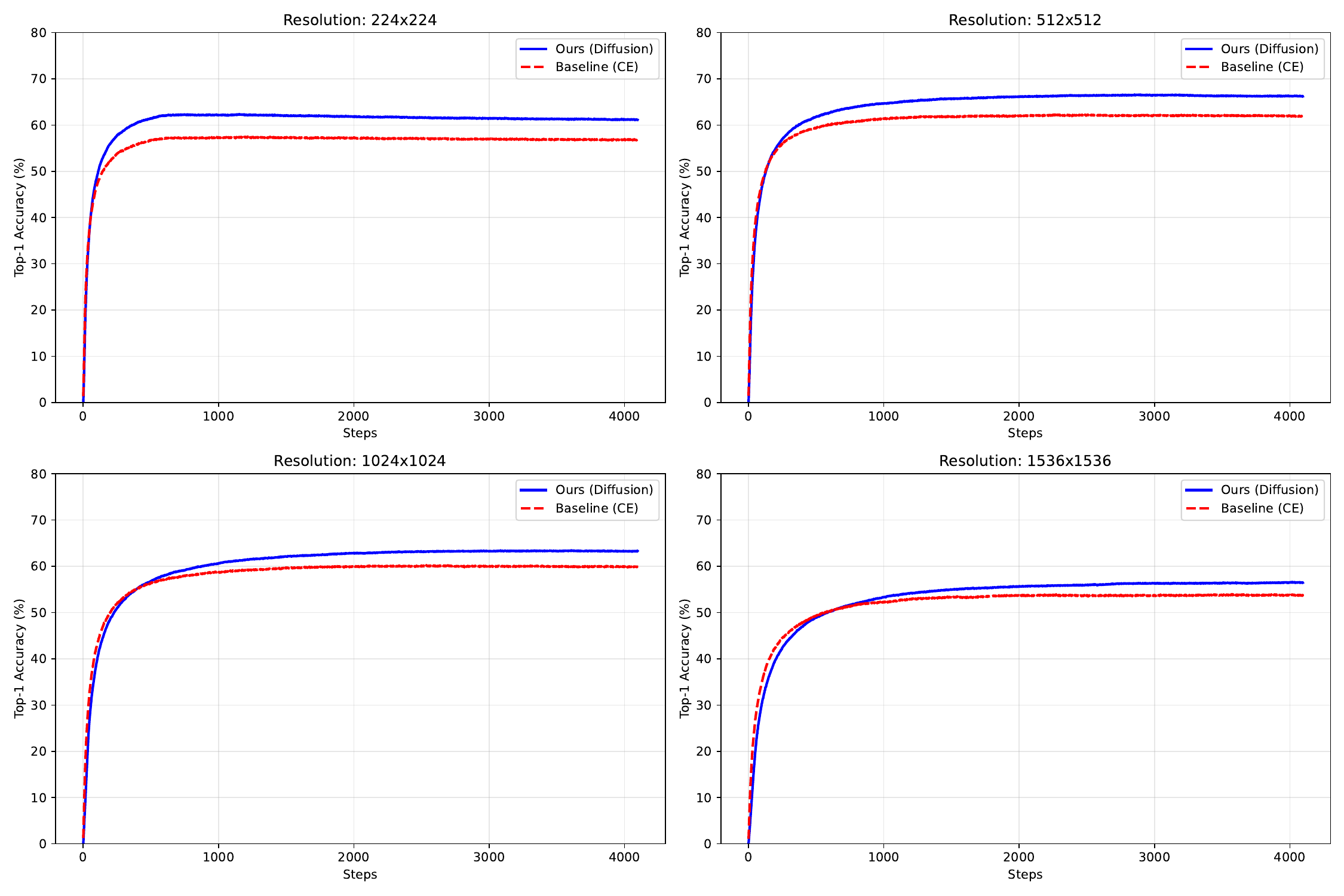}
    \caption{\textbf{Detailed Loss Ablation Comparison.} Step-wise top-1 accuracy comparison between MambaEye trained with our Diffusion-inspired loss (Blue) and the standard Cross-Entropy loss (Red dashed). Our method consistently achieves higher accuracy and faster convergence across all resolutions, particularly at higher resolutions where the dense supervision signal is most critical.}
    \label{fig:loss_ablation_suppl}
\end{figure*}

To further validate the effectiveness of our diffusion-inspired loss, we provide a detailed step-wise comparison against the standard Cross-Entropy (CE) loss baseline. Figure~\ref{fig:loss_ablation_suppl} illustrates the validation accuracy curves for both models across four different resolutions ($224^2, 512^2, 1024^2, 1536^2$) as a function of the number of steps processed.

\subsection{Comprehensive Performance Analysis}
\begin{figure*}[p]
    \centering
    \begin{subfigure}[b]{0.32\textwidth}
        \includegraphics[width=\linewidth]{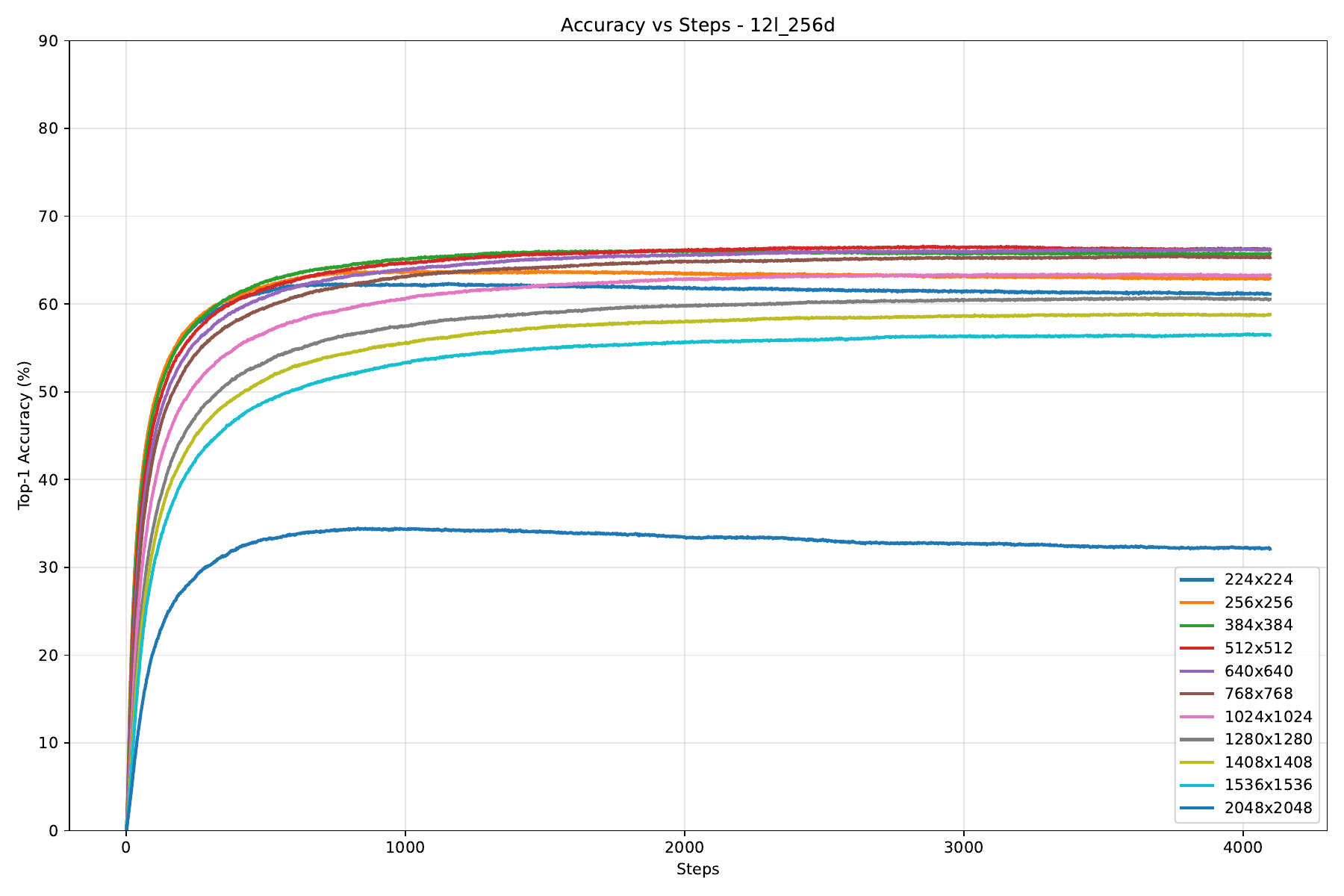}
        \caption{MambaEye-T (Pre-trained)}
    \end{subfigure}
    \hfill
    \begin{subfigure}[b]{0.32\textwidth}
        \includegraphics[width=\linewidth]{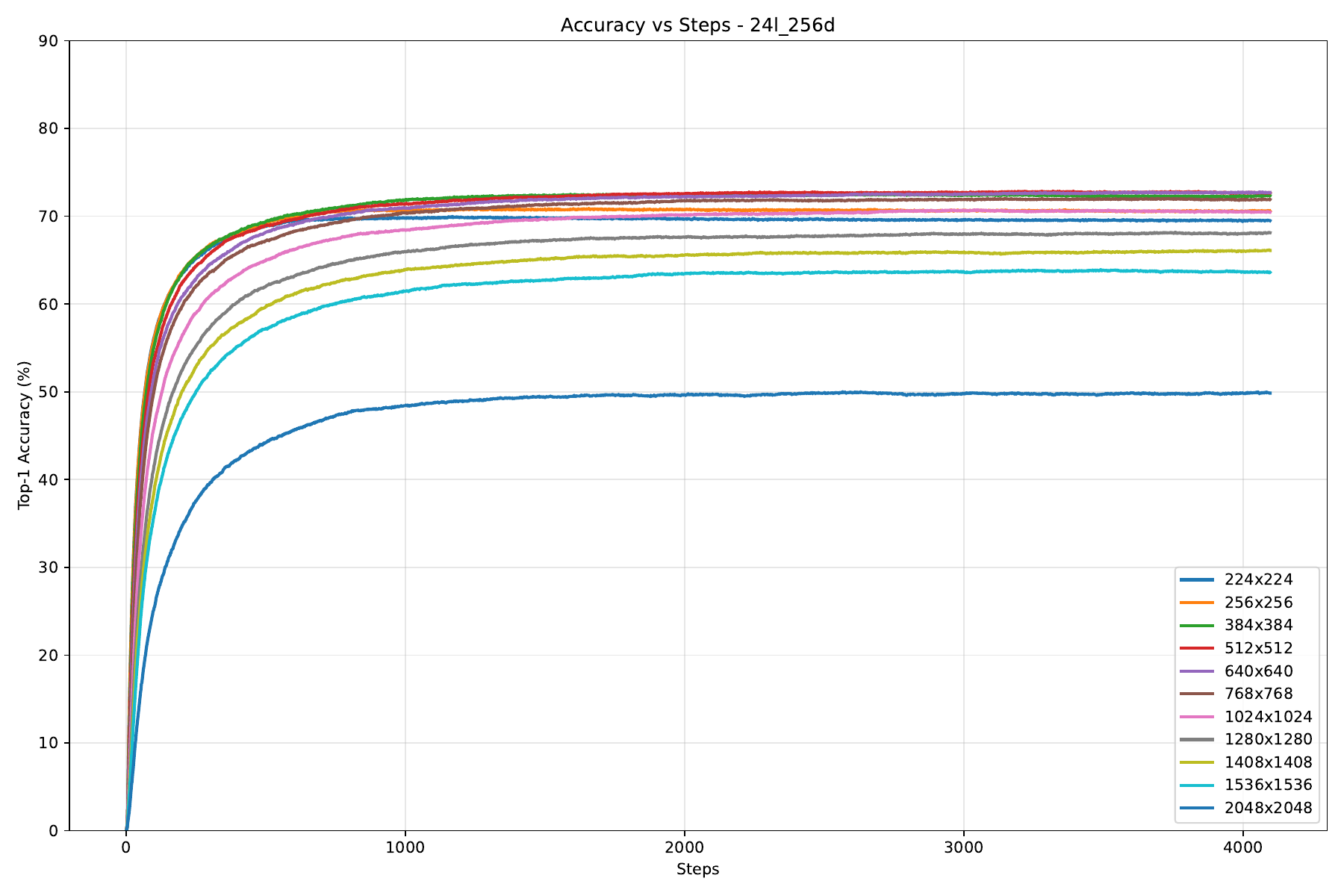}
        \caption{MambaEye-S (Pre-trained)}
    \end{subfigure}
    \hfill
    \begin{subfigure}[b]{0.32\textwidth}
        \includegraphics[width=\linewidth]{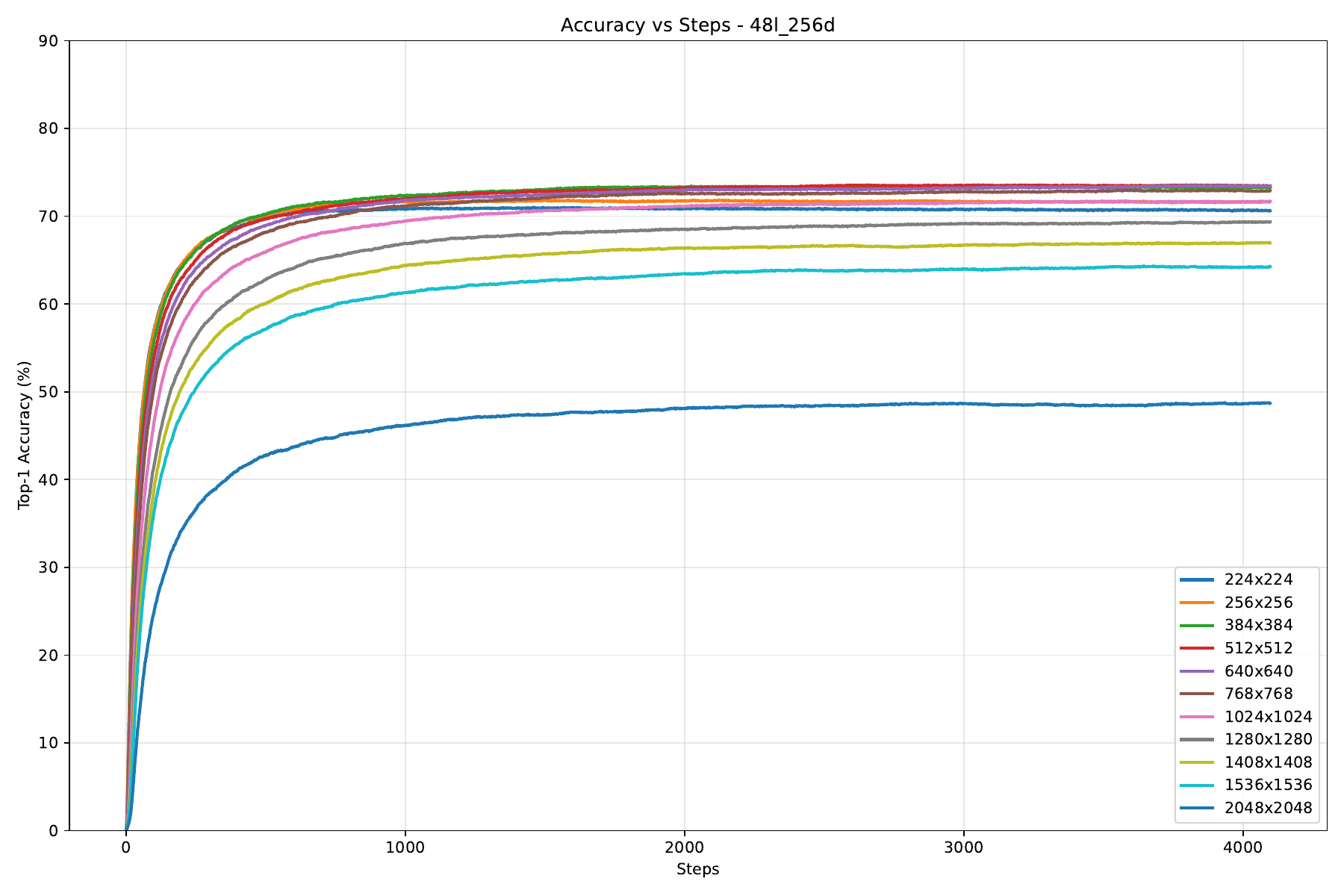}
        \caption{MambaEye-B (Pre-trained)}
    \end{subfigure}

    \vspace{1em}

    \begin{subfigure}[b]{0.32\textwidth}
        \includegraphics[width=\linewidth]{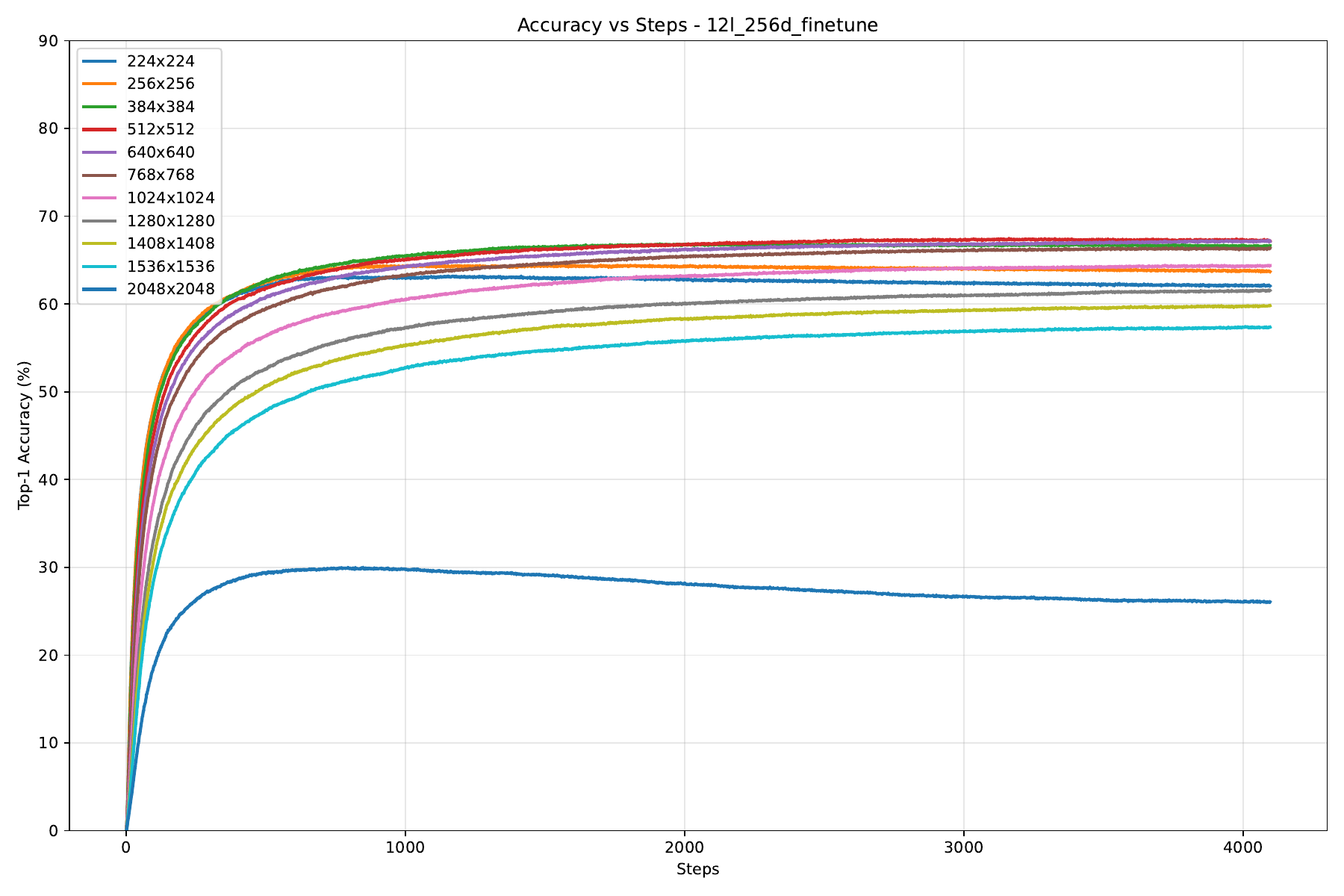}
        \caption{MambaEye-T (Fine-tuned)}
    \end{subfigure}
    \hfill
    \begin{subfigure}[b]{0.32\textwidth}
        \includegraphics[width=\linewidth]{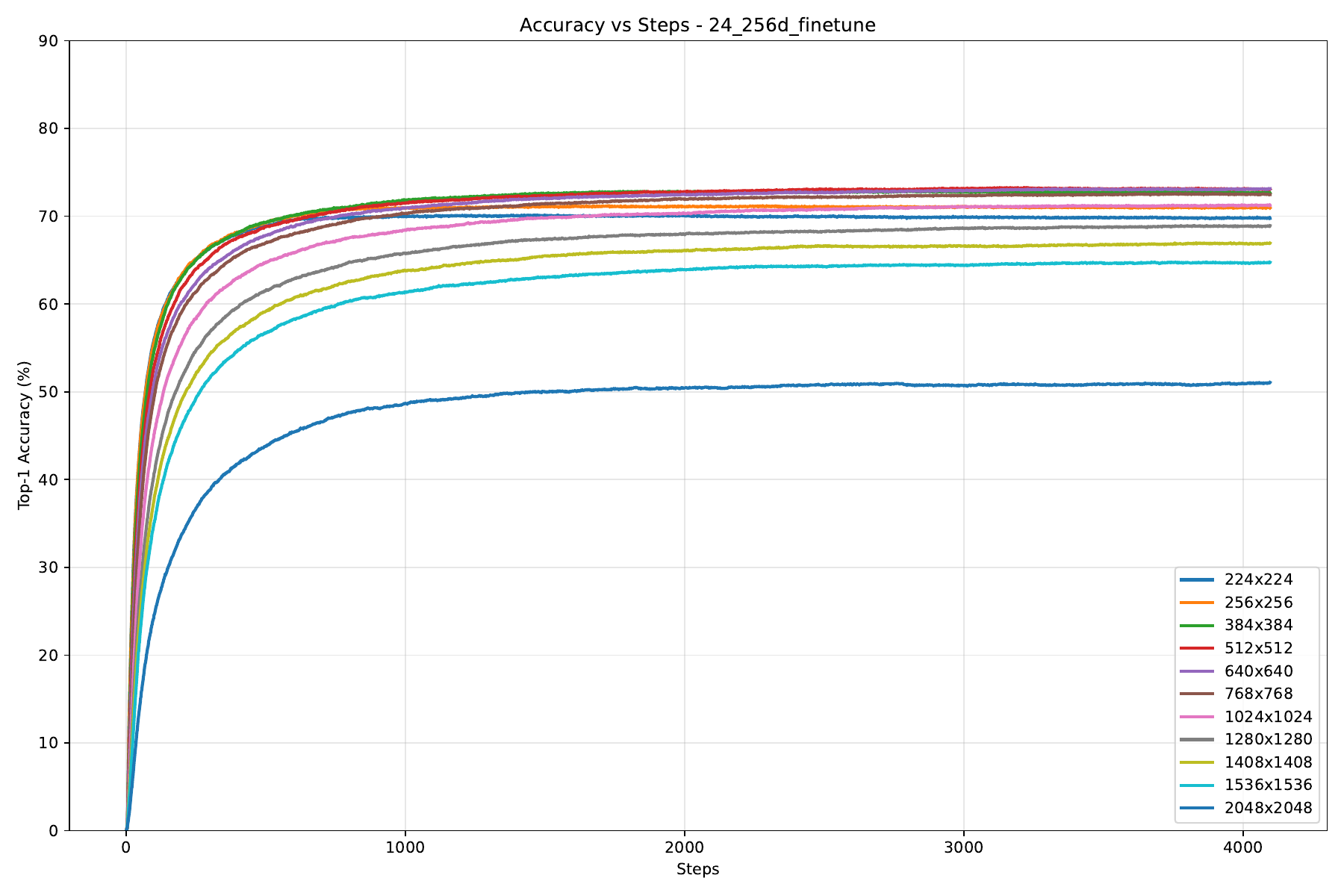}
        \caption{MambaEye-S (Fine-tuned)}
    \end{subfigure}
    \hfill
    \begin{subfigure}[b]{0.32\textwidth}
        \includegraphics[width=\linewidth]{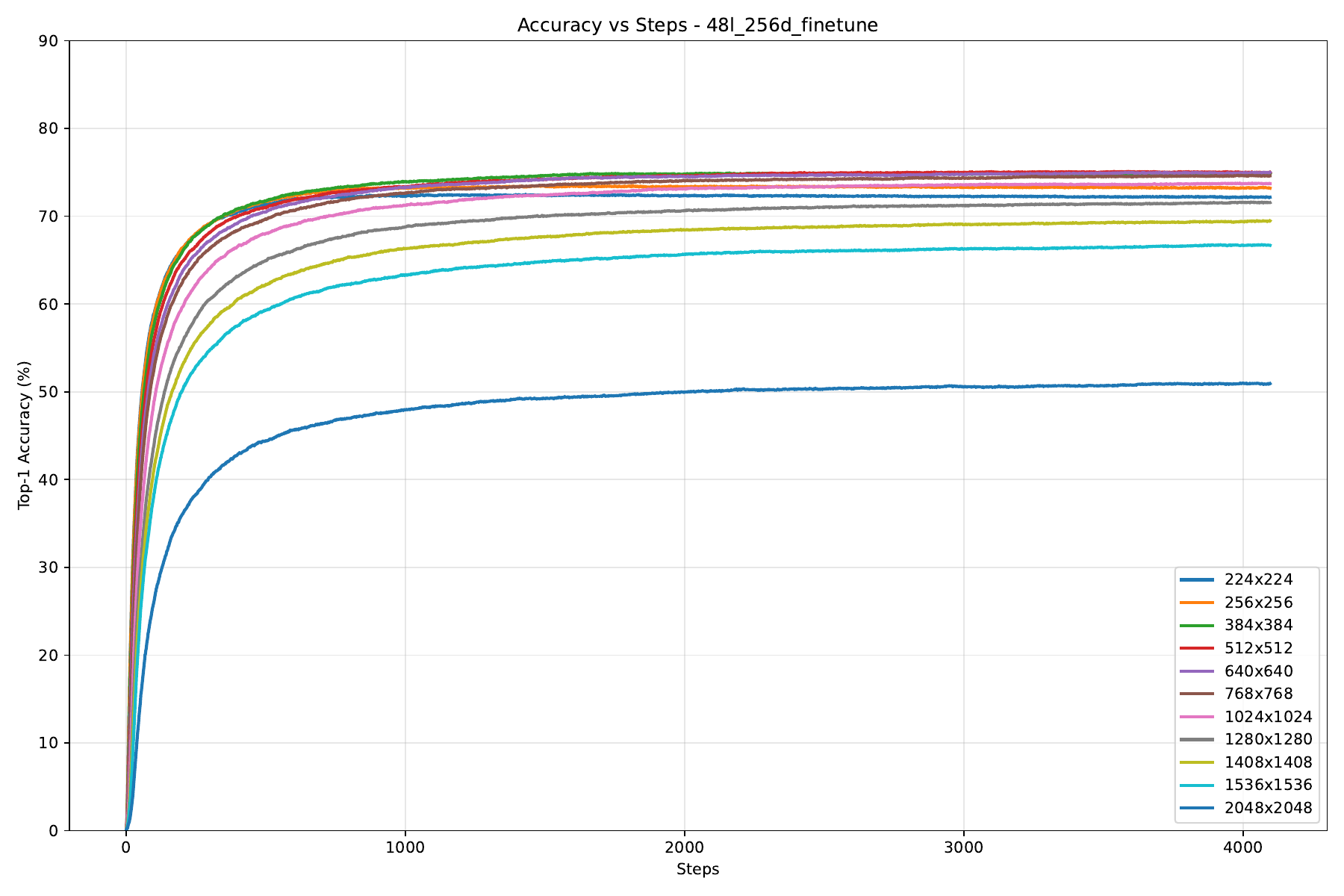}
        \caption{MambaEye-B (Fine-tuned)}
    \end{subfigure}

    \vspace{1em}

    \begin{subfigure}[b]{0.32\textwidth}
        \includegraphics[width=\linewidth]{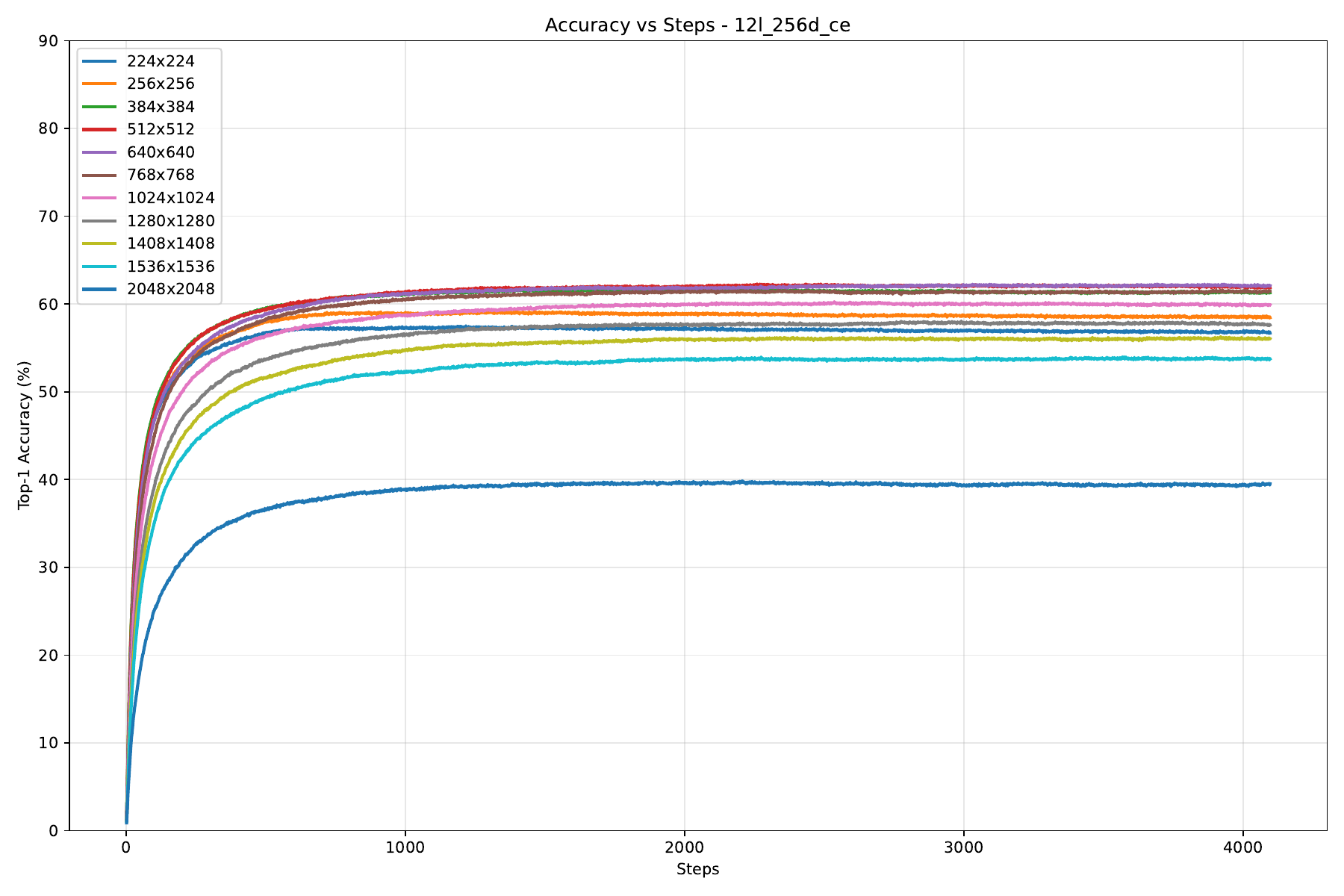}
        \caption{MambaEye-T (CE Loss Baseline)}
    \end{subfigure}
    \hfill
    \begin{subfigure}[b]{0.32\textwidth}
        \includegraphics[width=\linewidth]{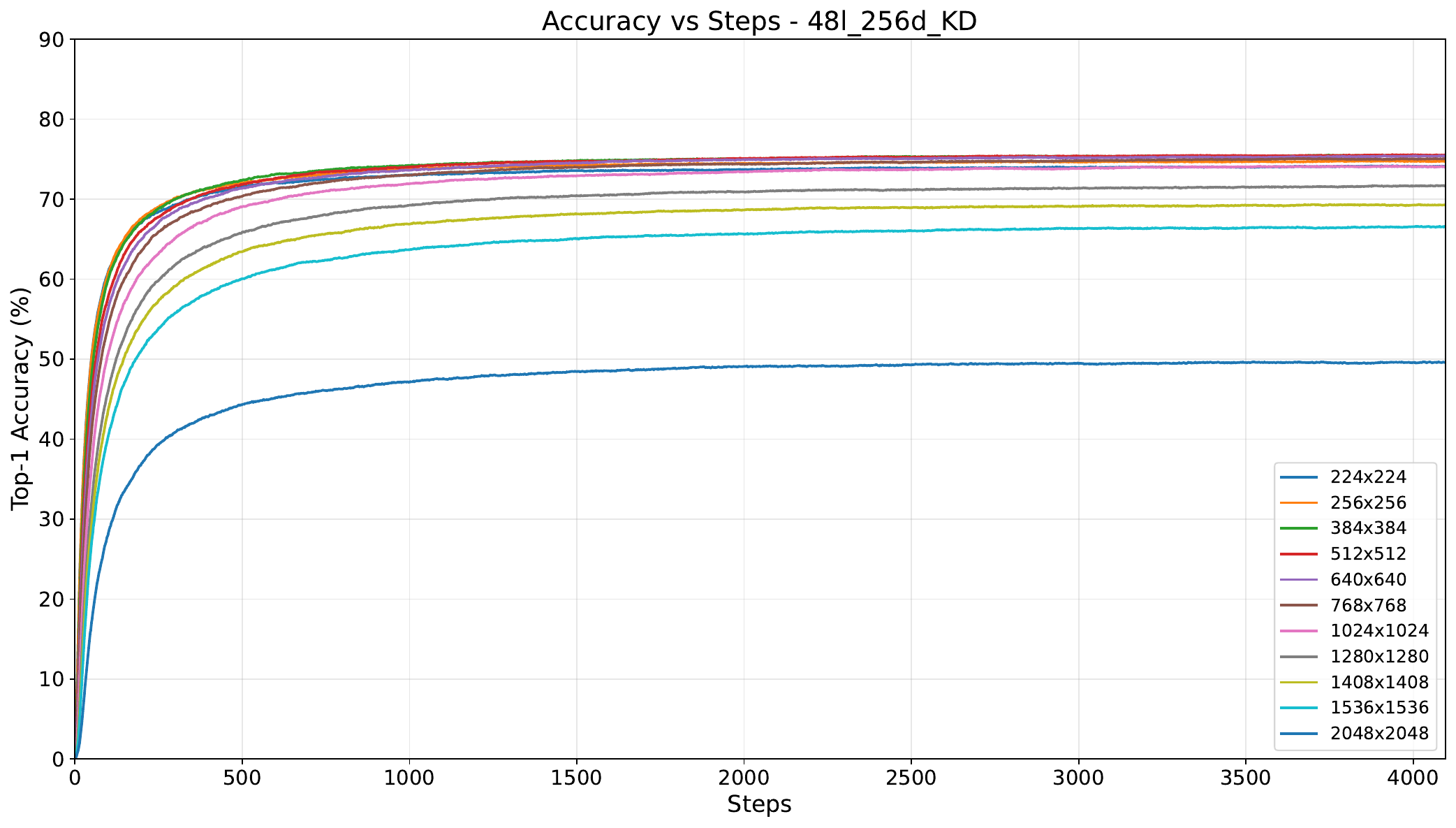}
        \caption{MambaEye-B (KD)}
    \end{subfigure}
    \caption{\textbf{Comprehensive Performance Curves.} Accuracy vs.\ Steps for all MambaEye variants across multiple resolutions. Rows 1--2: pre-trained and fine-tuned models. Row 3: knowledge distillation and CE loss ablation.}
    \label{fig:comprehensive_results}
\end{figure*}

We present the full performance curves for all trained MambaEye variants across different resolutions in Figure~\ref{fig:comprehensive_results}.

\subsection{Scanning Path Analysis}
\label{sec:suppl_scanning}

We provide a detailed analysis of the scanning path ablation presented in the main paper. The key finding is that scan order performance is governed by how quickly the path exposes \emph{global} image context to the causal model. We evaluated nine scanning patterns using MambaEye-B (FT) across resolutions from $224^{2}$ to $2048^{2}$.

\paragraph{Full Resolution Results.} Table~\ref{tab:scan_full} presents the complete results. The patterns fall into two clear groups: \emph{global-coverage} patterns (Golden, Random fixed grid) that maintain accuracy across resolutions, and \emph{local-traversal} patterns (Column, Snake, Spiral, Hilbert, Diagonal, Raster) that collapse at high resolutions.

\begin{table*}[h]
\centering
\caption{Full scanning path results on ImageNet-1K using MambaEye-B (FT), $T{=}4096$. Top-1 accuracy (\%).}
\label{tab:scan_full}
\footnotesize
\setlength{\tabcolsep}{4pt}
\begin{tabular}{@{}lcccccccccc@{}}
\toprule
Scanning Path & $224^{2}$ & $256^{2}$ & $384^{2}$ & $512^{2}$ & $640^{2}$ & $768^{2}$ & $1024^{2}$ & $1280^{2}$ & $1536^{2}$ \\
\midrule
Golden & 59.4 & 64.8 & 71.5 & 73.2 & 73.7 & 73.7 & 73.1 & 70.8 & \textbf{67.7} \\
Random (fixed grid) & \textbf{60.3} & \textbf{64.8} & \textbf{72.0} & \textbf{73.6} & \textbf{74.2} & \textbf{74.2} & \textbf{73.8} & \textbf{71.5} & 66.5  \\
\midrule
Diagonal & 51.2 & 53.5 & 54.7 & 51.4 & 47.0 & 39.8 & 26.0 & 14.2 & 9.4 \\
Spiral & 51.6 & 53.7 & 50.6 & 45.8 & 40.9 & 36.5 & 27.2 & 20.2 & 7.6 \\
H. Raster & 49.6 & 51.6 & 49.9 & 44.9 & 39.8 & 32.6 & 18.7 & 12.4 & 8.0 \\
Hilbert & 45.0 & 45.7 & 44.2 & 40.7 & 36.3 & 31.7 & 24.1 & 16.7 & 12.9  \\
Column Raster & 45.3 & 46.8 & 45.3 & 38.9 & 31.7 & 23.5 & 12.5 & 9.2 & 6.0\\
H. Snake & 43.2 & 43.1 & 37.5 & 31.5 & 27.4 & 22.8 & 13.8 & 9.6 & 6.4 \\
C. Snake & 42.4 & 43.1 & 40.1 & 35.6 & 31.4 & 25.2 & 14.0 & 9.3 & 6.0  \\
\bottomrule
\end{tabular}
\end{table*}

\paragraph{Local vs.\ Global Context.} Deterministic paths that traverse mostly adjacent patches collapse at high resolutions because they lack early global context. In contrast, the \textbf{Golden} scan jumps forward by $\approx 61.8\%$ of the total sequence length at every step, guaranteeing that sample points are maximally spread across the entire image from the start. This provides global context without randomness and achieves performance comparable to random scanning.

\begin{figure*}[p]
    \centering
    \begin{subfigure}[b]{0.32\textwidth}
        \includegraphics[width=\linewidth]{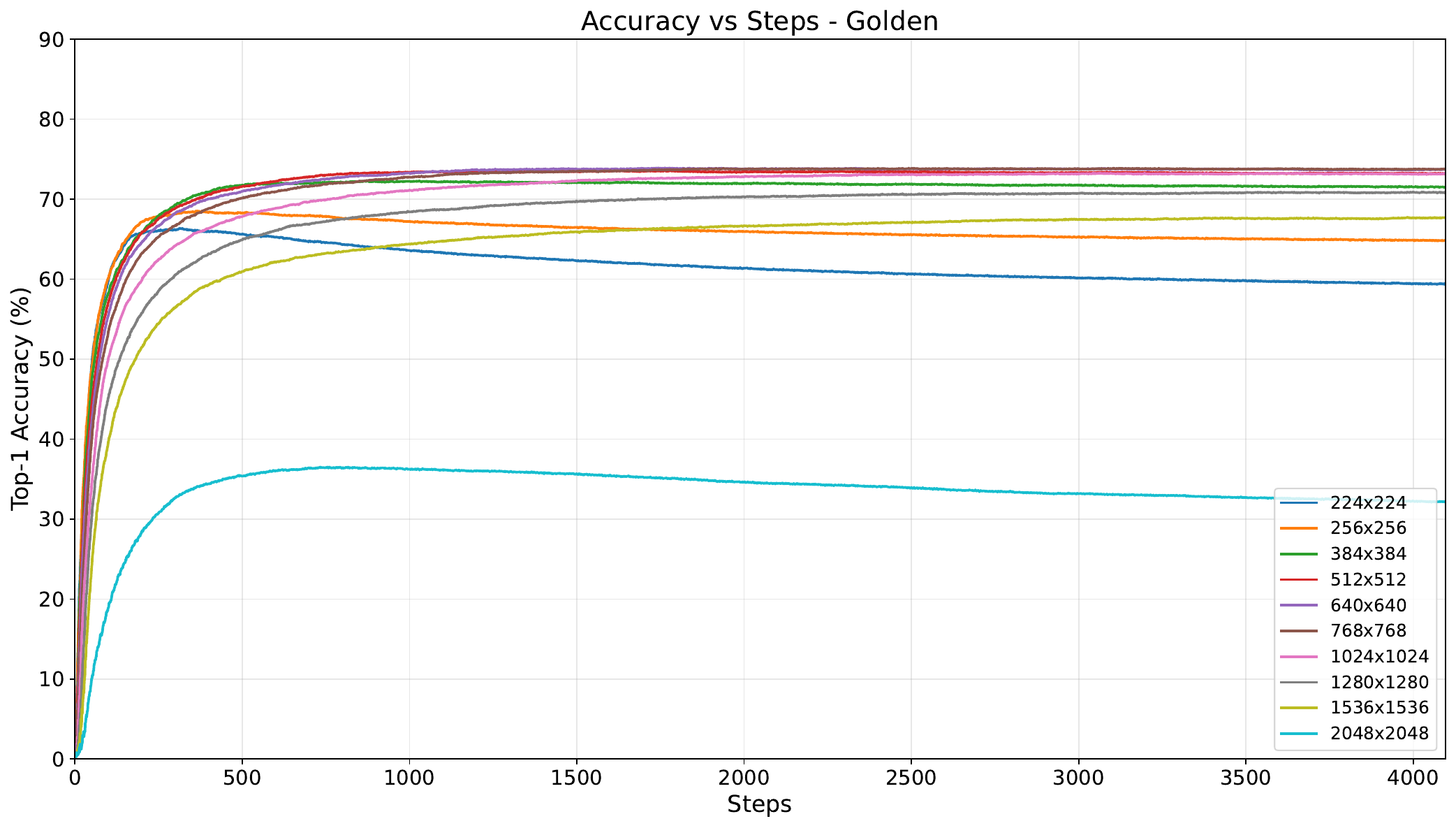}
        \caption{Golden}
    \end{subfigure}
    \hfill
    \begin{subfigure}[b]{0.32\textwidth}
        \includegraphics[width=\linewidth]{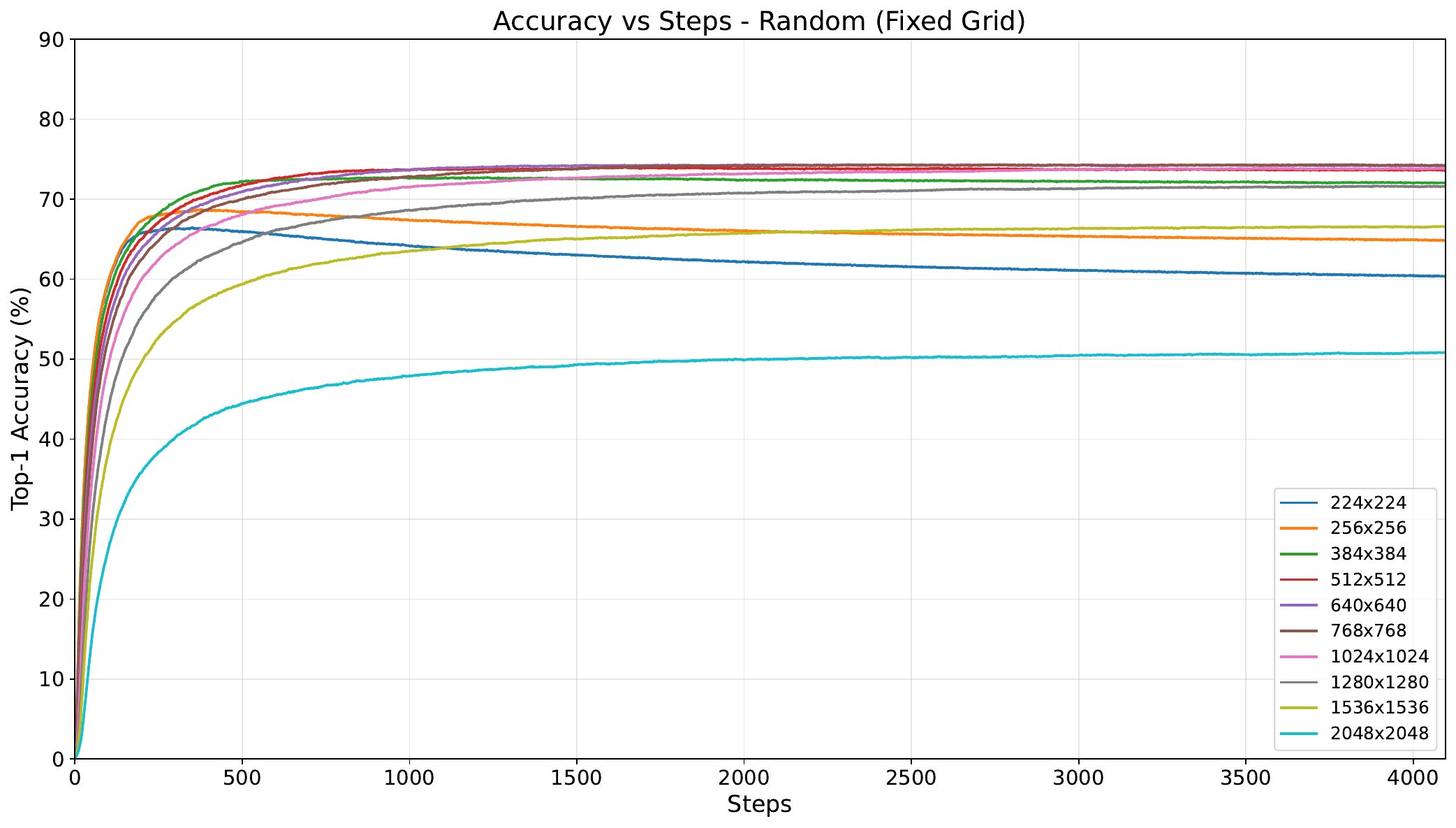}
        \caption{Random (Fixed Grid)}
    \end{subfigure}
    \hfill
    \begin{subfigure}[b]{0.32\textwidth}
        \includegraphics[width=\linewidth]{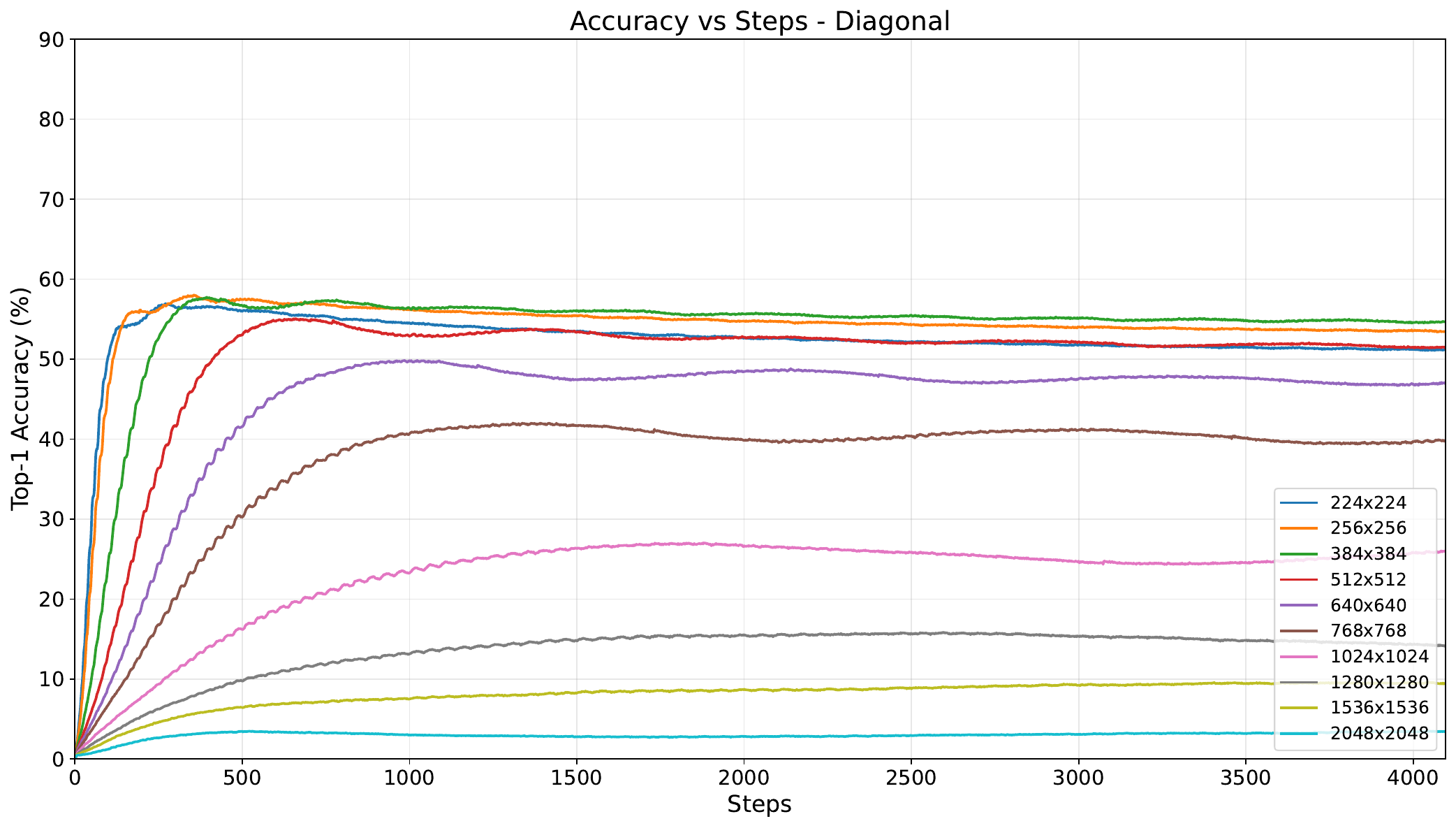}
        \caption{Diagonal}
    \end{subfigure}

    \vspace{1em}

    \begin{subfigure}[b]{0.32\textwidth}
        \includegraphics[width=\linewidth]{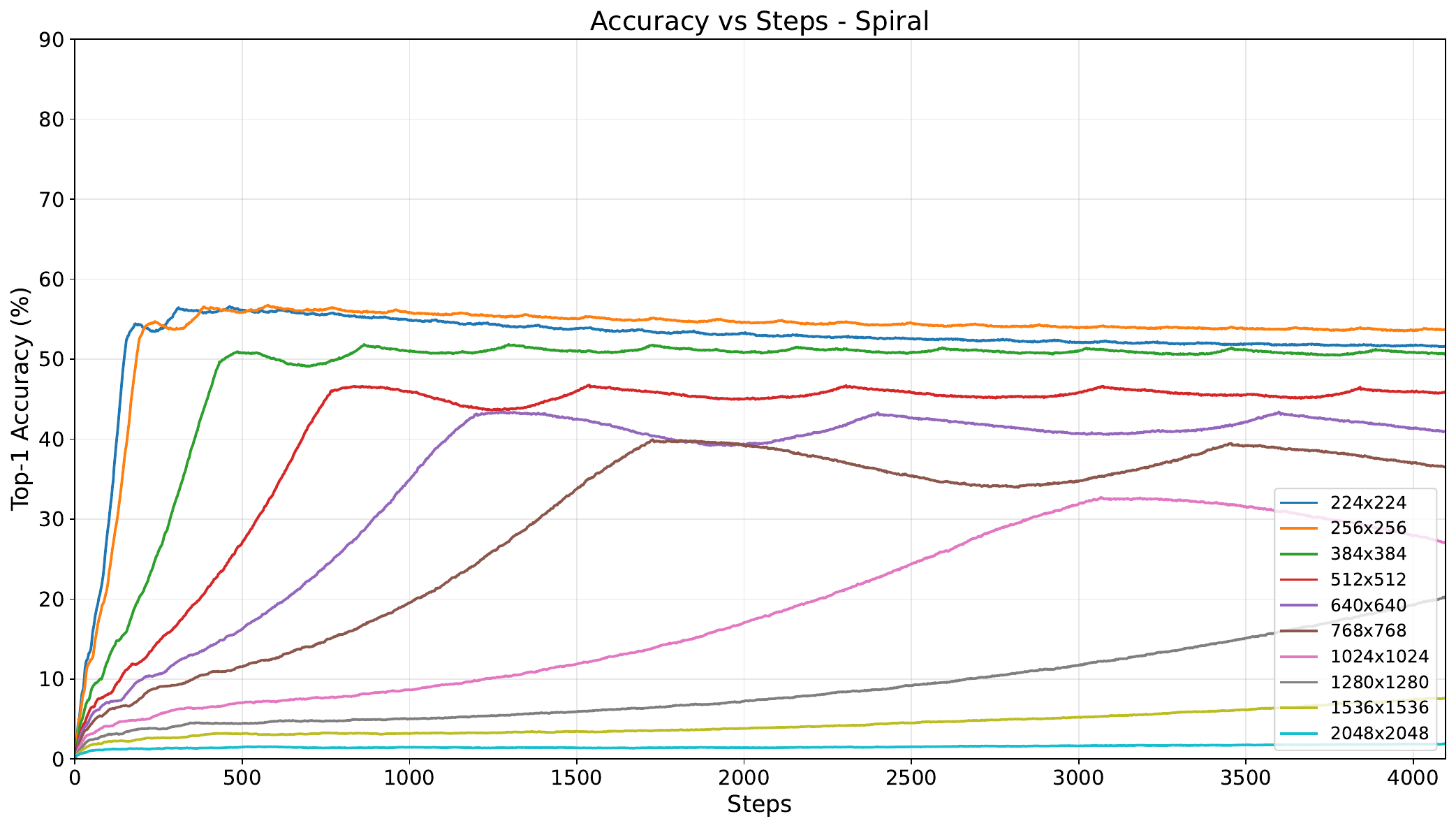}
        \caption{Spiral}
    \end{subfigure}
    \hfill
    \begin{subfigure}[b]{0.32\textwidth}
        \includegraphics[width=\linewidth]{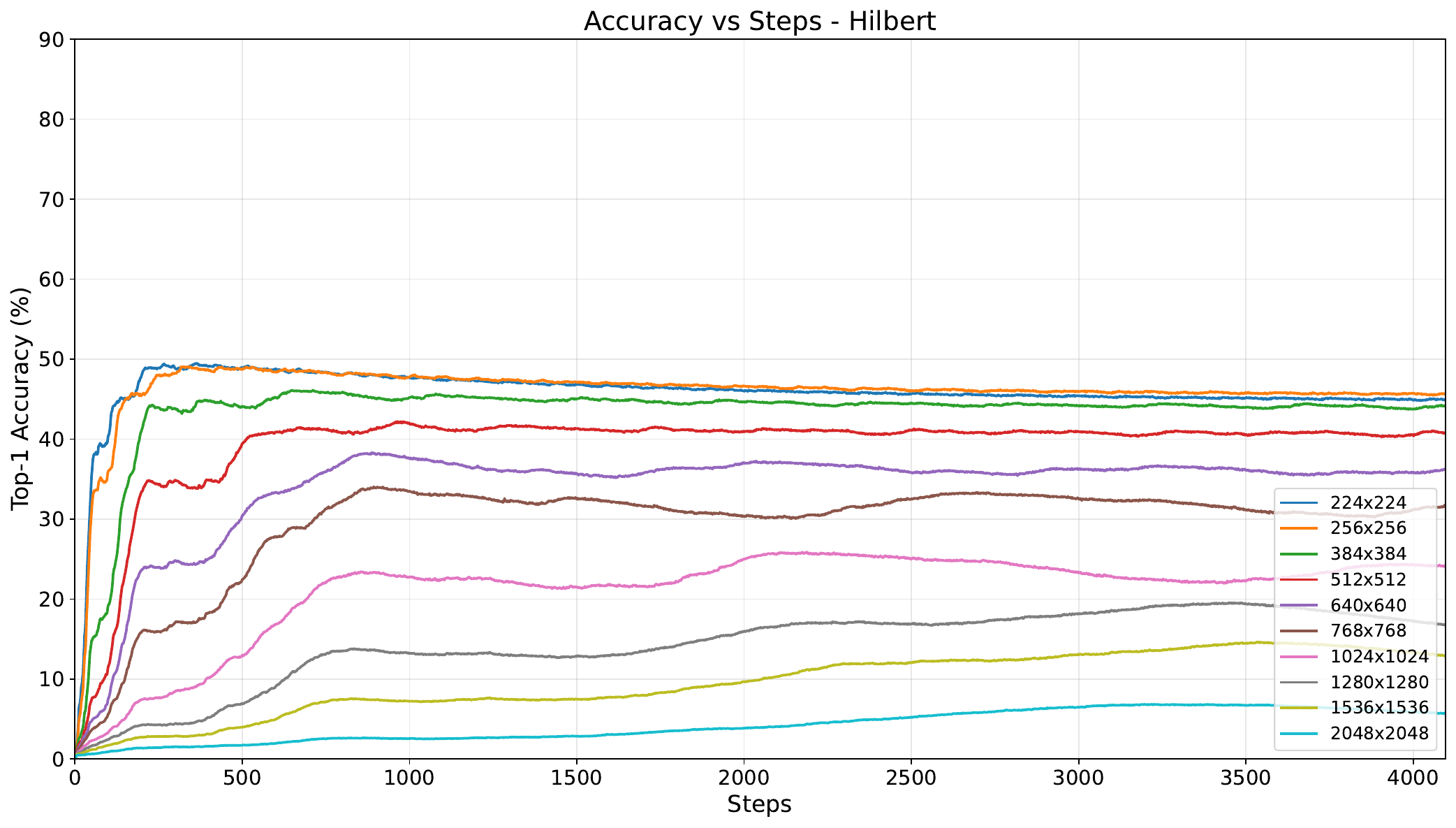}
        \caption{Hilbert}
    \end{subfigure}
    \hfill
    \begin{subfigure}[b]{0.32\textwidth}
        \includegraphics[width=\linewidth]{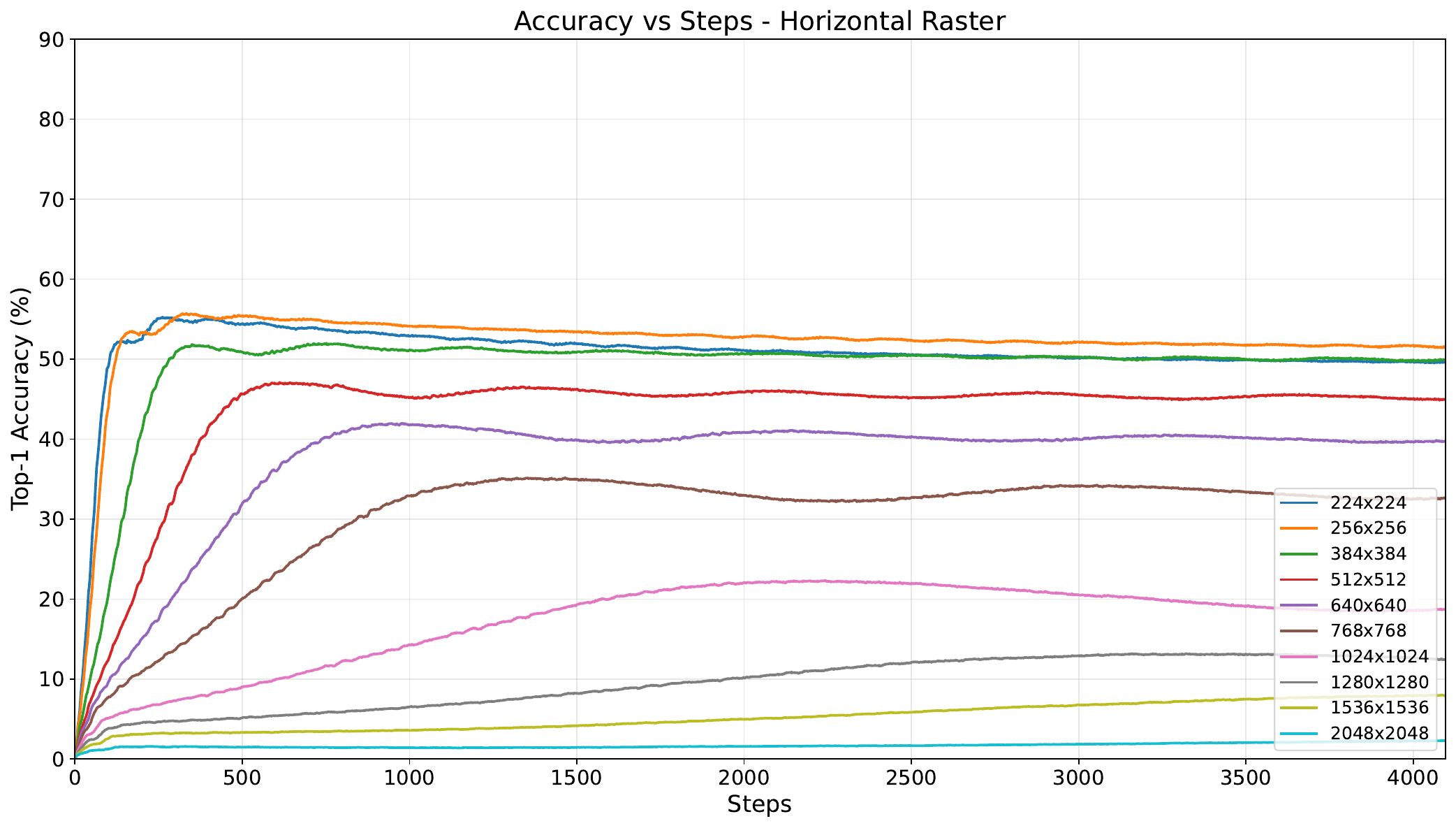}
        \caption{H. Raster}
    \end{subfigure}

    \vspace{1em}

    \begin{subfigure}[b]{0.32\textwidth}
        \includegraphics[width=\linewidth]{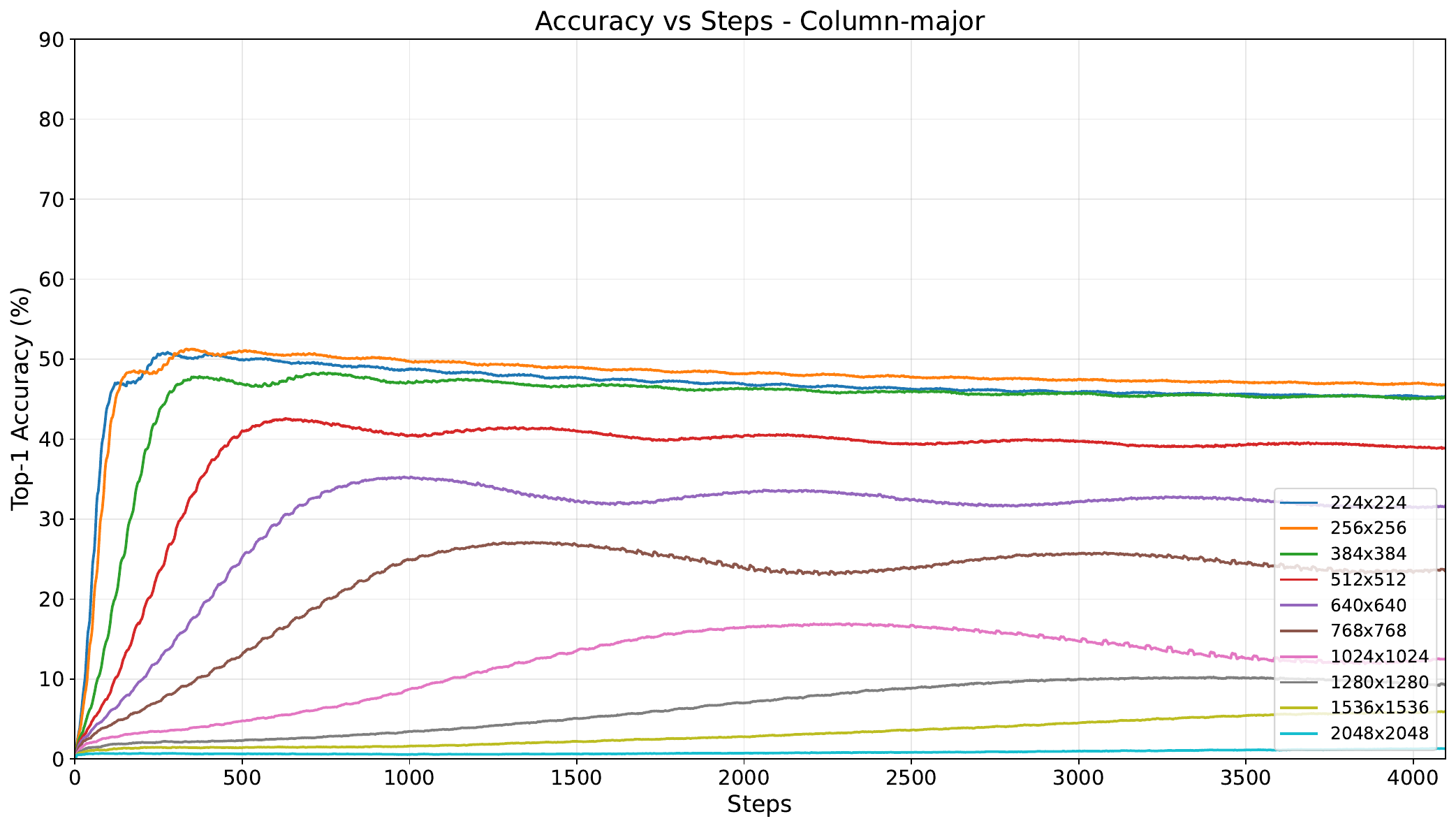}
        \caption{Column Raster}
    \end{subfigure}
    \hfill
    \begin{subfigure}[b]{0.32\textwidth}
        \includegraphics[width=\linewidth]{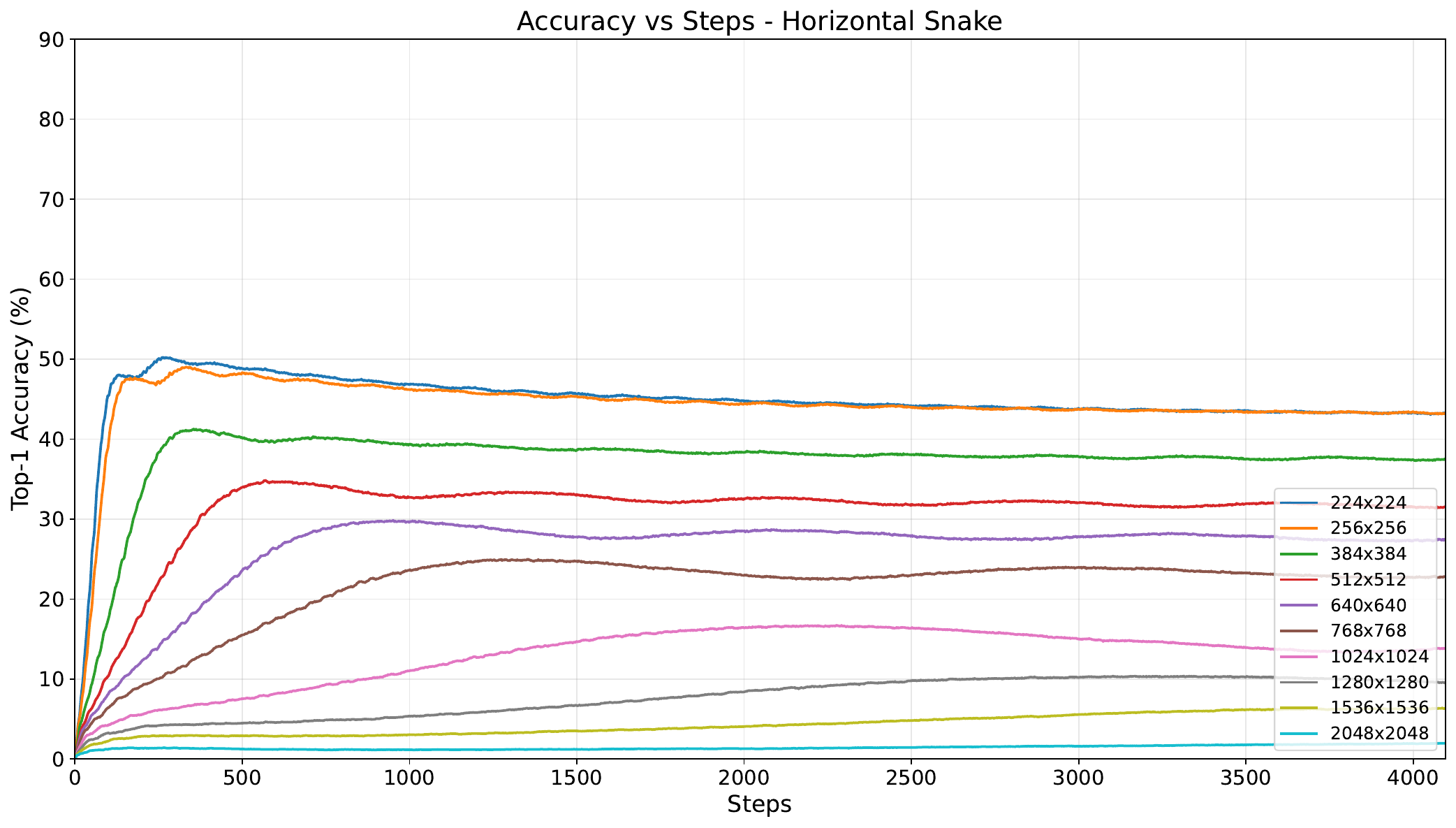}
        \caption{H. Snake}
    \end{subfigure}
    \hfill
    \begin{subfigure}[b]{0.32\textwidth}
        \includegraphics[width=\linewidth]{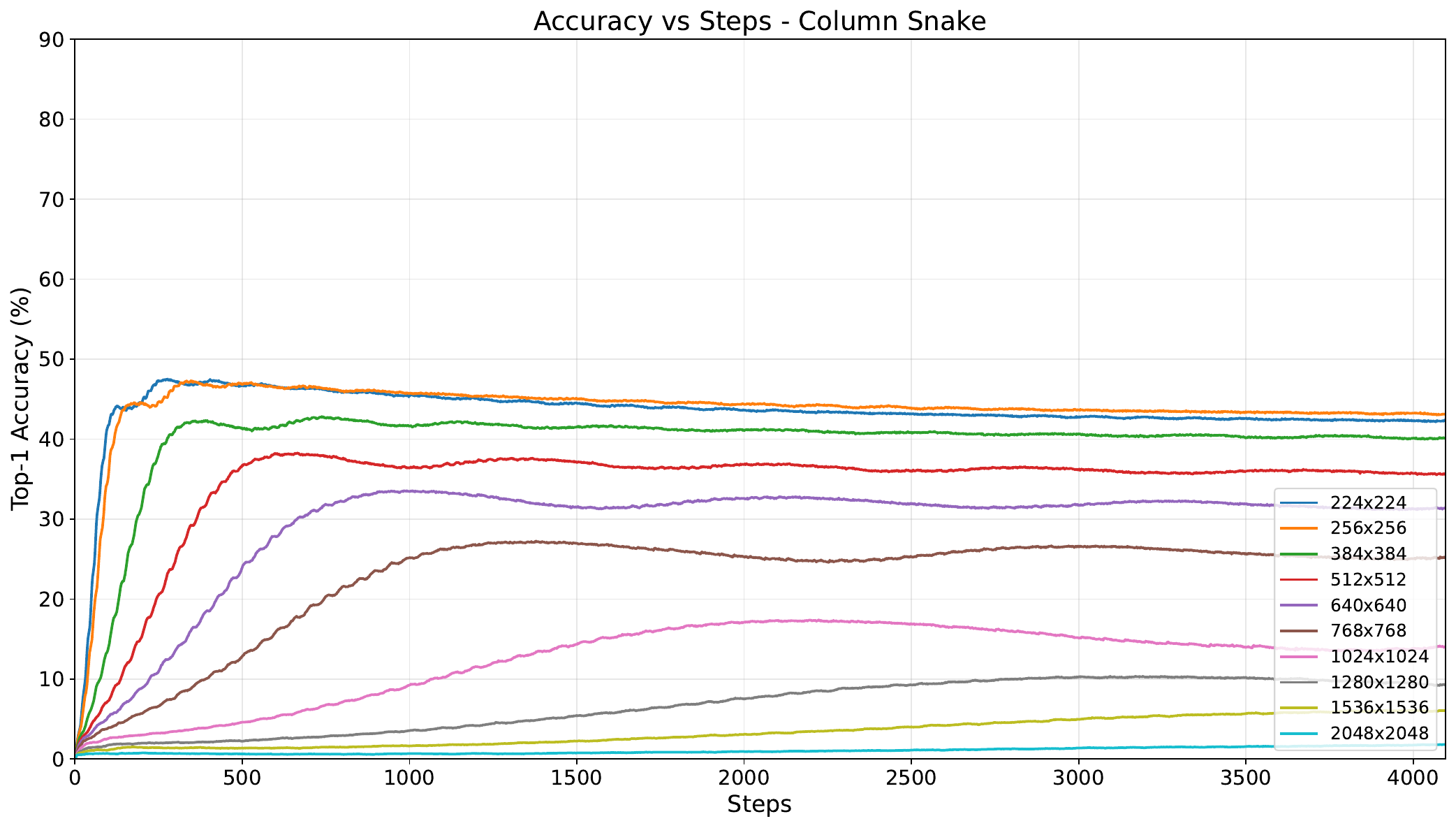}
        \caption{C. Snake}
    \end{subfigure}

    \caption{\textbf{Step-wise accuracy for all scanning patterns.} Each subplot shows accuracy vs.\ sequence length ($T{=}4096$) for MambaEye-B (FT) across all resolutions. Global-coverage patterns (a--b) produce smooth, saturating curves. Local-traversal patterns (c--i) exhibit oscillations and degradation at high resolutions.}
    \label{fig:scan_individual}
\end{figure*}

\paragraph{Per-Pattern Step-Wise Analysis.}
Figure~\ref{fig:scan_individual} shows the step-wise accuracy curves for each scanning pattern. Several notable behaviors emerge:
\begin{itemize}
    \item \textbf{Golden and Random (fixed grid)} exhibit smooth, monotonically increasing curves resembling our default random sampling (cf.\ Figure~\ref{fig:comprehensive_results}), confirming that early global coverage is the key ingredient.
    \item \textbf{Column Raster and Horizontal Raster} show pronounced periodic oscillations caused by row/column boundary transitions, where large spatial jumps disrupt the causal state.
    \item \textbf{Snake variants} (Horizontal and Column) suppress the high-frequency oscillations by ensuring spatially continuous transitions at row/column boundaries, but still degrade because they only visit nearby patches in succession.
    \item \textbf{Diagonal} achieves the highest peak among local patterns at lower resolutions ($54.7\%$ at $384^{2}$) because its traversal covers more spatial extent per step than row/column-based scans, but this advantage disappears at high resolutions.
    \item \textbf{Hilbert} curve, despite being a space-filling fractal designed for locality preservation, performs poorly because it still restricts the model to a local neighborhood for the first portion of the sequence.
    \item \textbf{Spiral} shows an interesting pattern: accuracy increases as the spiral expands outward from the center, providing gradually increasing spatial coverage, but collapses once the spiral wraps and re-enters previously visited regions.
\end{itemize}

\begin{figure*}[p]
    \centering
    \begin{subfigure}[b]{0.48\textwidth}
        \includegraphics[width=\linewidth]{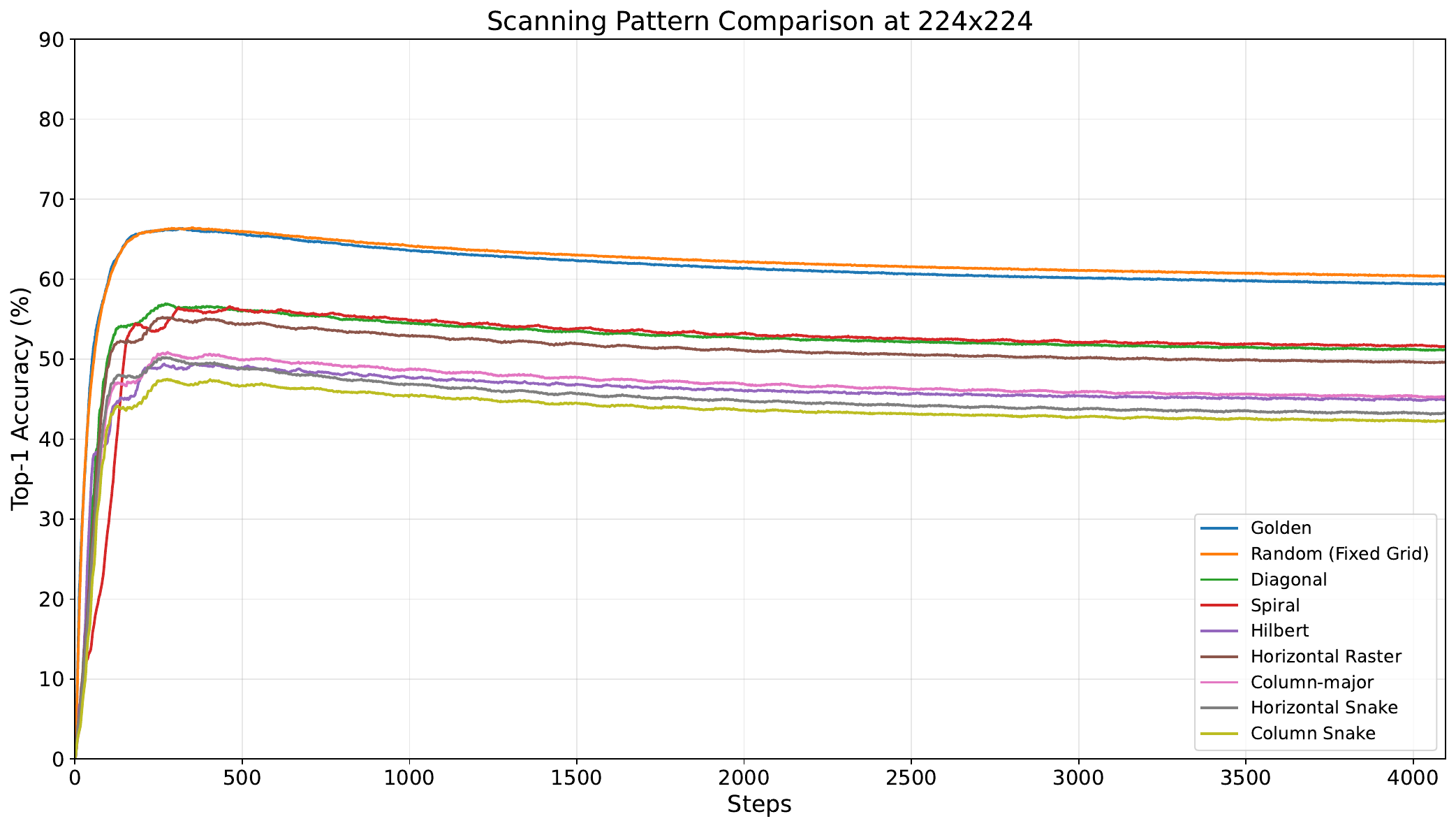}
        \caption{$224^{2}$}
    \end{subfigure}
    \hfill
    \begin{subfigure}[b]{0.48\textwidth}
        \includegraphics[width=\linewidth]{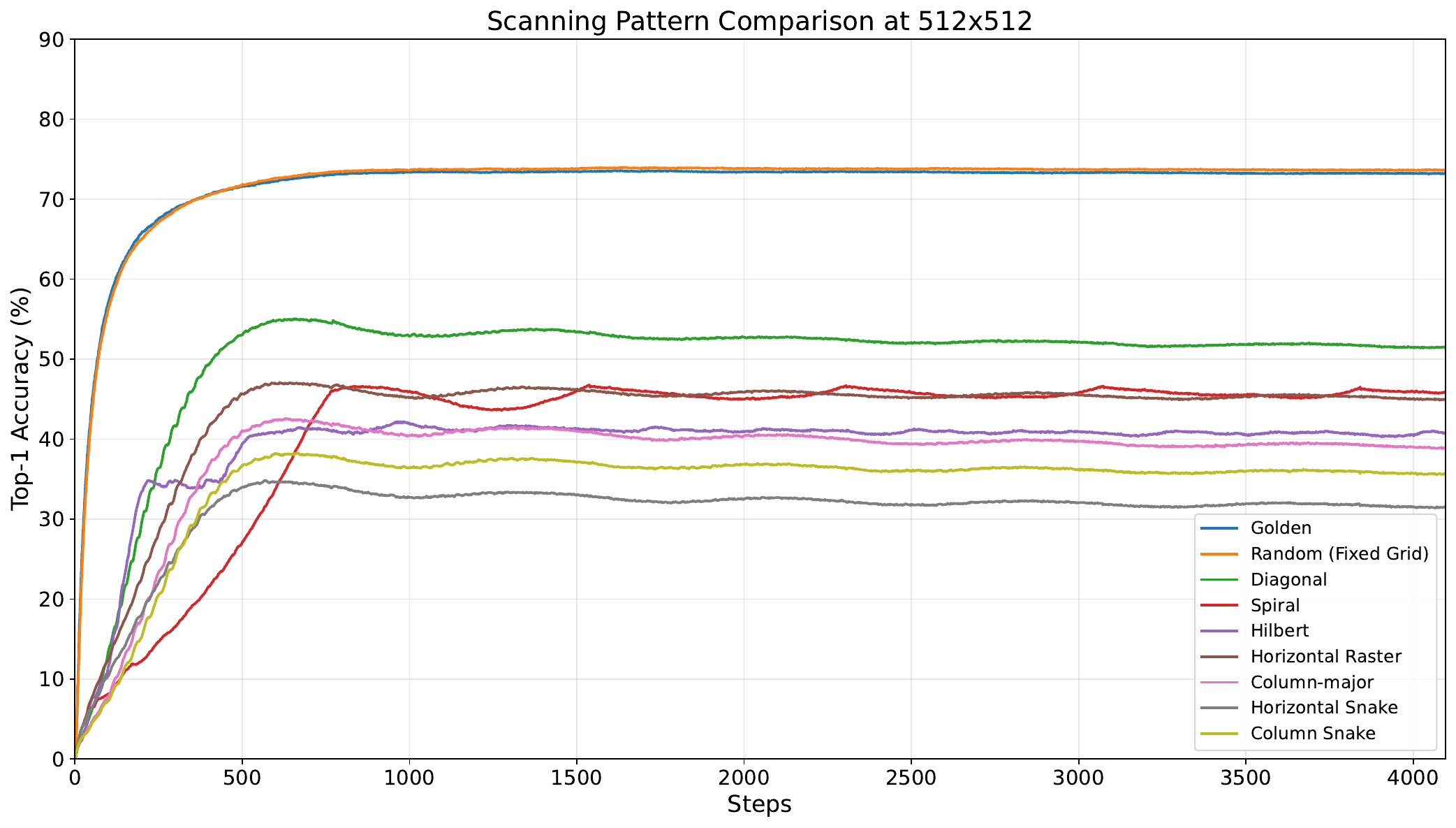}
        \caption{$512^{2}$}
    \end{subfigure}

    \vspace{1em}

    \begin{subfigure}[b]{0.48\textwidth}
        \includegraphics[width=\linewidth]{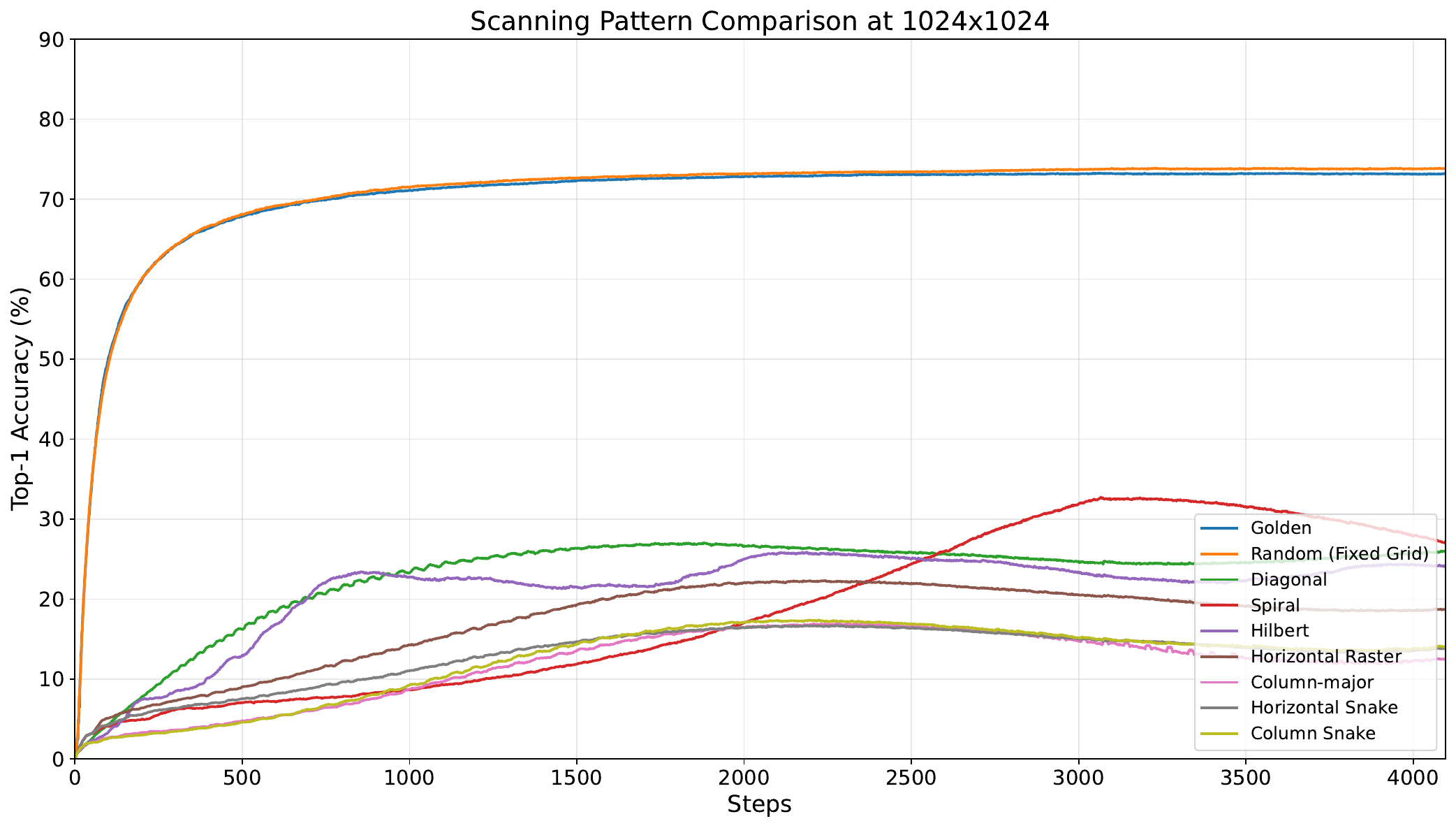}
        \caption{$1024^{2}$}
    \end{subfigure}
    \hfill
    \begin{subfigure}[b]{0.48\textwidth}
        \includegraphics[width=\linewidth]{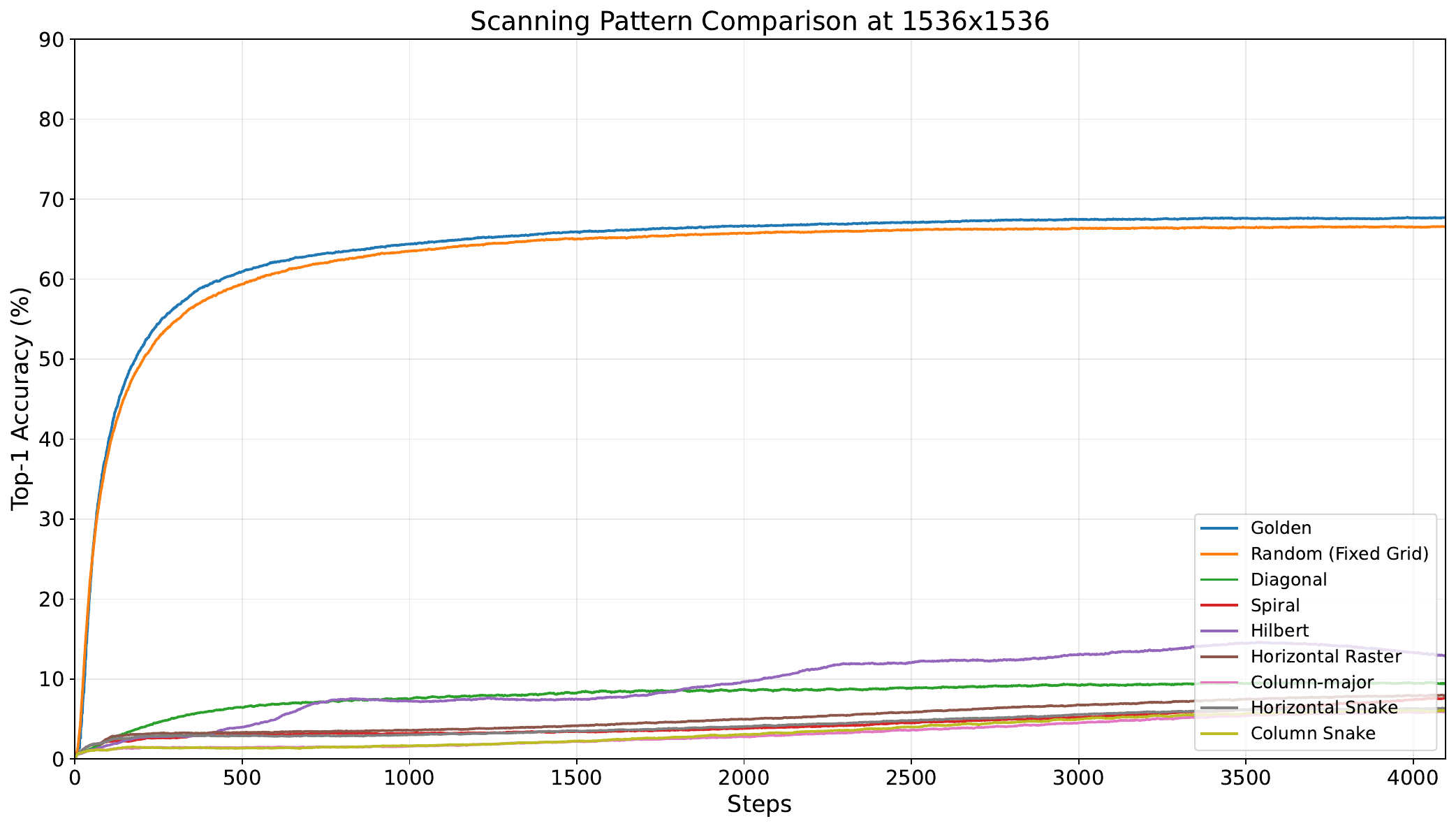}
        \caption{$1536^{2}$}
    \end{subfigure}

    \caption{\textbf{Cross-pattern comparison at key resolutions.} All nine scanning patterns compared at $224^{2}$, $512^{2}$, $1024^{2}$, and $1536^{2}$. Global-coverage patterns (Golden, Random fixed grid) consistently outperform local-traversal patterns, with the gap widening at higher resolutions.}
    \label{fig:scan_comparison}
\end{figure*}

\paragraph{Cross-Pattern Comparison.}
Figure~\ref{fig:scan_comparison} directly compares all patterns at four key resolutions. At $512^{2}$, Golden and Random (fixed grid) reach ${\sim}73\%$ while local patterns plateau between $31$--$51\%$. The gap widens dramatically at $1536^{2}$, where global-coverage patterns maintain $66$--$68\%$ accuracy while all local patterns fall below $13\%$.

\subsection{Knowledge Distillation: Full Resolution Breakdown}
\label{sec:suppl_kd}

Table~\ref{tab:kd_full} presents the complete knowledge distillation results across all evaluated resolutions. The step-wise accuracy progression is shown in Figure~\ref{fig:comprehensive_results}(g).

\begin{table}[h]
\centering
\caption{Full KD results on ImageNet-1K ($T{=}4096$). Teacher: DeiT-III-Base (86M, 83.8\%). Best per resolution in \textbf{bold}.}
\label{tab:kd_full}
\small
\setlength{\tabcolsep}{3pt}
\begin{tabular}{@{}lcccccccccc@{}}
\toprule
Model & $224^{2}$ & $256^{2}$ & $384^{2}$ & $512^{2}$ & $640^{2}$ & $768^{2}$ & $1024^{2}$ & $1280^{2}$ & $1408^{2}$ & $1536^{2}$ \\
\midrule
Base & 70.6 & 71.7 & 73.3 & 73.5 & 73.4 & 72.9 & 71.7 & 69.4 & 67.0 & 64.3\\
Base (FT) & 72.2 & 73.2 & 74.8 & 75.0 & 75.0 & 74.6 & 73.7 & 71.5 & \textbf{69.5} & \textbf{66.7} \\
Base (KD) & \textbf{74.1} & \textbf{74.8} & \textbf{75.5} & \textbf{75.5} & \textbf{75.4} & \textbf{75.0} & \textbf{74.1} & \textbf{71.7} & 69.3 & 66.5 \\
\bottomrule
\end{tabular}
\end{table}

The KD model achieves the highest accuracy from $224^{2}$ through $1024^{2}$, with gains of $+0.4$ to $+1.9$ over the fine-tuned model. Only at $1408^{2}$ and $1536^{2}$ does the fine-tuned model pull slightly ahead ($+0.2$ points), as it was explicitly trained on longer sequences ($T{=}2048$). The KD model, trained only at $T{=}1024$, generalizes well to $4\times$ longer evaluation sequences. The teacher's knowledge transfers across resolutions: DeiT-III processes images only at $224^{2}$, yet the student benefits at all resolutions, enabled by the information-ratio-weighted KD loss.

\end{document}